\PassOptionsToPackage{numbers,sort}{natbib}
\documentclass{article}
\usepackage[final,main]{neurips_2026}
\usepackage[utf8]{inputenc}
\usepackage[T1]{fontenc}
\usepackage{url}
\usepackage{booktabs}
\usepackage{amsfonts}
\usepackage{amsmath}
\usepackage{amssymb}
\usepackage{nicefrac}
\usepackage{microtype}
\usepackage{xcolor}
\usepackage{graphicx}
\usepackage{multirow}
\usepackage{makecell}
\usepackage{algorithm}
\usepackage{algpseudocode}
\setcitestyle{numbers,square,comma}
\usepackage{enumitem}
\usepackage{adjustbox}
\usepackage{wrapfig}
\usepackage{needspace}
\usepackage{caption}
\usepackage[most]{tcolorbox}
\usepackage[most]{tcolorbox}
\tcbuselibrary{skins,breakable,listings}
\usepackage{xcolor}

\definecolor{RepositoryPink}{RGB}{220,60,140}
\definecolor{PromptBlue}{RGB}{38,103,166}
\definecolor{PromptDarkBlue}{RGB}{25,72,120}
\definecolor{PromptLightBlue}{RGB}{246,250,255}

\definecolor{PromptOrange}{RGB}{194,105,35}
\definecolor{PromptDarkOrange}{RGB}{150,72,18}
\definecolor{PromptLightOrange}{RGB}{255,250,244}

\newtcblisting{bluepromptbox}[2][]{
  enhanced,
  breakable,
  listing only,
  listing engine=listings,
  colback=PromptLightBlue,
  colframe=PromptBlue,
  colbacktitle=PromptDarkBlue,
  coltitle=white,
  boxrule=0.7pt,
  arc=1.5mm,
  left=2mm,
  right=2mm,
  top=2mm,
  bottom=2mm,
  title={#2},
  fonttitle=\bfseries\footnotesize\sffamily,
  attach boxed title to top left={xshift=2mm,yshift=-2mm},
  boxed title style={
    colback=PromptDarkBlue,
    colframe=PromptDarkBlue,
    arc=1mm,
    boxrule=0pt,
    left=2mm,
    right=2mm,
    top=0.6mm,
    bottom=0.6mm
  },
  before skip=8pt,
  after skip=8pt,
  listing options={
    basicstyle=\ttfamily\scriptsize,
    breaklines=true,
    breakatwhitespace=false,
    columns=fullflexible,
    keepspaces=true,
    showstringspaces=false
  },
  #1
}

\newtcblisting{orangepromptbox}[2][]{
  enhanced,
  breakable,
  listing only,
  listing engine=listings,
  colback=PromptLightOrange,
  colframe=PromptOrange,
  colbacktitle=PromptDarkOrange,
  coltitle=white,
  boxrule=0.7pt,
  arc=1.5mm,
  left=2mm,
  right=2mm,
  top=2mm,
  bottom=2mm,
  title={#2},
  fonttitle=\bfseries\footnotesize\sffamily,
  attach boxed title to top left={xshift=2mm,yshift=-2mm},
  boxed title style={
    colback=PromptDarkOrange,
    colframe=PromptDarkOrange,
    arc=1mm,
    boxrule=0pt,
    left=2mm,
    right=2mm,
    top=0.6mm,
    bottom=0.6mm
  },
  before skip=8pt,
  after skip=8pt,
  listing options={
    basicstyle=\ttfamily\scriptsize,
    breaklines=true,
    breakatwhitespace=false,
    columns=fullflexible,
    keepspaces=true,
    showstringspaces=false
  },
  #1
}
\usepackage[
    colorlinks=true,
    linkcolor=black,
    citecolor=black,
    urlcolor=black
]{hyperref}
\title{Train the Agent, Not the Expert: \\
	 Learning to Harness Heterogeneous Experts for Multi-Turn Visual Reasoning}

\author{%
  Yaowu Fan\textsuperscript{1,}\thanks{\looseness=-1 \textls[-15]{Equal contribution.\enspace\textsuperscript{$\dagger$}Corresponding author. Andy Jin-Hua Ma is also with Guangdong Provincial Key Laboratory of Information Security Technology and Key Laboratory of Machine Intelligence and Advanced Computing, Ministry of Education, China.}}\quad
  Tao Han\textsuperscript{2,$\ast$}\quad
  Dazhao Du\textsuperscript{2}\quad
  Andy Jin-Hua Ma\textsuperscript{1,$\dagger$}\quad
  Jia Wan\textsuperscript{3} \\
  \textsuperscript{1}Sun Yat-sen University\quad
  \textsuperscript{2}HKUST\quad
  \textsuperscript{3}Harbin Institute of Technology (Shenzhen) \\
  {\footnotesize\texttt{\{fywyukee,hantao10200,dudazhao16,jiawan1998\}@gmail.com, majh8@mail.sysu.edu.cn}}
}

\makeatletter
\renewcommand{\section}{%
  \@startsection{section}{1}{\z@}{-5pt}{4pt}{\large\bfseries\raggedright}}
\renewcommand{\subsection}{%
  \@startsection{subsection}{2}{\z@}{-4pt}{3pt}{\normalsize\bfseries\raggedright}}
\renewcommand{\subsubsection}{%
  \@startsection{subsubsection}{3}{\z@}{-4pt}{3pt}{\normalsize\bfseries\raggedright}}
\makeatother
\begin{document}
\raggedbottom

\maketitle

\begin{abstract}
Recent advances in computer vision have produced powerful specialized models for detection, segmentation, counting, and other visual tasks.
However, these models are typically optimized for isolated tasks, limiting their ability to support general-purpose visual intelligence, particularly when complex language understanding and dense small-object perception are required.
We propose \textbf{VisHarness}, a trainable visual agent that decouples high-level perception, reasoning, and decision-making from low-level task execution.
Rather than training a task-specific model, we train VisHarness to harness a set of carefully designed heterogeneous visual experts.
This paradigm preserves the agent's general intelligence while leveraging the precision of specialized visual experts for specific tasks.
With only lightweight training, VisHarness learns a generalizable policy for harnessing visual experts, enabling it to solve fundamental vision tasks in diverse complex scenarios through multi-turn interactions with these experts.
To enable efficient on-policy reinforcement learning in a live environment, we introduce dynamic visual memory archiving to reduce the visual-token overhead that rapidly accumulates during multi-turn interactions with visual experts.
Experiments on four representative benchmarks spanning reasoning segmentation, generalized referring expression segmentation, dense small-object detection, and referring expression counting show that VisHarness substantially outperforms existing general-purpose models and is competitive with or superior to task-specific models.
Our code, model weights, and training data are publicly available at {\hypersetup{urlcolor=RepositoryPink}\url{https://github.com/fyw1999/VisHarness}}.

\end{abstract}
\section{Introduction}

\begin{wrapfigure}[13]{r}{0.52\textwidth}
  \vspace{\dimexpr-12pt-1.5\baselineskip\relax}
  \centering
  \captionsetup{skip=4pt}
  \includegraphics[width=0.96\linewidth]{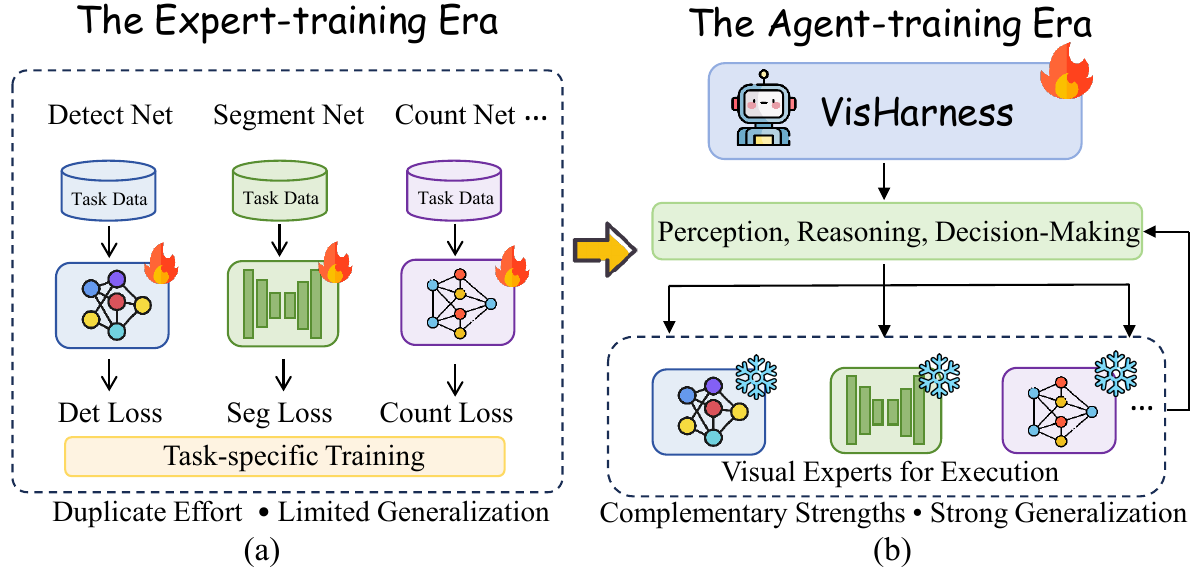}
  \caption{%
    \textbf{From expert training to expert harnessing.}
    (a): Traditional vision methods train a separate specialist for each sub-task.
    (b): VisHarness learns a harnessing policy over a set of heterogeneous experts and thus can solve complex visual tasks through multi-turn interaction.}
    \label{fig:teaser}
\end{wrapfigure}

The past decade has witnessed rapid advances in computer vision. Powerful specialized models have emerged across multiple subfields, such as object detection \cite{YOLO26}, semantic segmentation \cite{SAM}, and object counting \cite{ZSOC}, achieving substantial progress on their respective tasks. 
The continuous improvement of these task-specific capabilities has driven computer vision beyond traditional perception tasks toward more open and complex visual understanding scenarios \cite{liu2023gres, dai2024rec8k}.

Despite their impressive performance, specialized visual models typically operate as isolated task solvers, each designed for a specific visual task with a fixed input-output format. For example, detection and segmentation models \cite{SAM, SAM3, ravi2024sam2, Deformable-DETR} excel at object localization and pixel-level mask generation, but often lack the ability to understand complex instructions. Counting models \cite{wan2021generalized, SDNet} effectively estimate dense object distributions
but provide limited support for instance-level reasoning. Visual grounding models \cite{liu2024grounding, wang2025refdetector} can interpret complex language descriptions, but often struggle with densely packed small objects. As a result, a single model is often insufficient to handle
complex visual requests (see Fig. \ref{fig:teaser}(a)). Real-world visual tasks often require combining multiple capabilities, such as first understanding complex expressions, then detecting or segmenting densely packed small objects, and finally verifying the outputs or reasoning
based on intermediate results.

In recent years, multimodal large language models (MLLMs) \cite{kimi2026k25, qwen3.5} have demonstrated strong vision-language understanding and reasoning capabilities, offering new possibilities for general-purpose visual intelligence. However, MLLMs themselves are not always well suited to precise visual tasks, such as segmentation, dense object detection, and accurate counting. Reliable performance on these tasks often requires task-specific architectural modifications
and large amounts of high-quality training data \cite{ jiang2026rexomni, GiT, lai2024lisa, tang2025ufo}.
Recently, some works \cite{RSAgent, SAM-Veteran, IBISAgent, SegAgent} enable MLLMs to interact with promptable segmentation models \cite{SAM} over multiple turns to iteratively refine target masks. However, these approaches remain focused on segmentation, leaving other visual tasks that require detecting and counting densely packed small objects largely unexplored.

To address the above issues, we propose \textbf{VisHarness}, a trainable high-level agent for harnessing heterogeneous visual experts. With only a small amount of training data, VisHarness can learn a policy for harnessing a carefully designed set of heterogeneous visual expert models, enabling it to solve a range of fundamental vision tasks. The core design principle of VisHarness is to decouple high-level perception, reasoning, and decision-making from low-level vision task execution. Specifically, the agent learns general perception, reasoning, and decision-making abilities, while delegating the execution of specific vision tasks to specialized visual experts. This design preserves the general intelligence of the agent while fully leveraging the precision advantages of specialized visual experts in concrete visual tasks (see Fig. \ref{fig:teaser}(b)). 
Accordingly, VisHarness can naturally support a variety of fundamental vision tasks, including generalized referring expression segmentation (GRES), reasoning segmentation, referring expression counting, and dense small-objects detection. Through multi-turn interactions with visual experts, VisHarness can not only handle complex visual tasks beyond the capabilities of a single expert but also identify and correct errors from previous turns. At each turn, VisHarness first observes the current image and text content, then reasons to make a decision. After a specific visual expert returns the execution result in visual and textual form, VisHarness can continue reasoning based on this feedback and decide in the next turn whether to refine the current result or invoke another visual expert, until the task is completed.

To learn a generalizable visual expert-harnessing policy, we train VisHarness with on-policy reinforcement learning (RL) in a live environment with deployed visual experts. However, visual feedback accumulates over multiple interaction turns, introducing substantial memory and context overhead that makes RL training difficult. To this end, we further introduce dynamic visual memory archiving, which dynamically archives outdated visual memories at each reasoning step and retains only the visual results produced by the latest interaction with a visual expert. This mechanism preserves necessary historical information while substantially reducing visual context overhead, enabling efficient trajectory sampling for on-policy RL. Since dynamic visual memory archiving changes the historical memory at each turn, we decompose the multi-turn Markov decision process into multiple single-turn Markov decision processes to maintain consistency between training and inference during the stage of RL training. In summary, our main contributions are as follows:
\begin{itemize}[leftmargin=*]
\setlength{\itemsep}{1pt}
\setlength{\topsep}{2pt}
\item We introduce VisHarness, a trainable visual agent that decouples high-level visual perception, reasoning, and decision-making from low-level visual task execution. It requires only a small amount of training data to learn a generalizable policy over a set of heterogeneous visual experts and can solve various complex vision tasks and self-correct through multi-turn reasoning.

\item We propose a dynamic visual memory archiving mechanism that substantially reduces the visual overhead in multi-turn interactions without compromising historical memory coherence and task performance. This enables efficient on-policy reinforcement learning of VisHarness in a live environment, allowing it to autonomously explore effective and generalizable policies under the guidance of trajectory-level rewards.

\item Experimental results on four benchmarks covering generalized referring expression segmentation (GRES), reasoning segmentation, referring expression counting, and  dense small-object detection show that VisHarness outperforms general-purpose models and performs  comparably to or better than specialized models. Strong performance on tasks and datasets unseen during training further demonstrates the generalizability of the expert harnessing policy learned by VisHarness.
\end{itemize}

\section{Related Work}
\label{sec:related}

\subsection{MLLM-Based Task-Specific Visual Models}
Recent works have adapted MLLMs into specialized visual models~\cite{lai2024lisa, wei2025instructseg, Text4Seg++, jiang2026rexomni}. LISA~\cite{lai2024lisa} introduces a special \texttt{<SEG>} token and combines its projected hidden representation with image features in a segmentation decoder to generate the target mask for a complex implicit query.
 Text4Seg++~\cite{Text4Seg++} extends the MLLM vocabulary with 126 semantic brick tokens, each encoding a run of foreground or background cells, enabling the model to autoregressively generate bounding boxes and mask sequences for fine-grained, language-guided segmentation without an additional segmentation decoder. Rex-Omni~\cite{jiang2026rexomni} repurposes the last 1,000 tokens in Qwen2.5-VL’s vocabulary as dedicated tokens for quantized relative coordinates, enabling the MLLM to autoregressively generate points, bounding boxes, and polygons to unify open-vocabulary detection, fine-grained referring expression grounding, OCR, and keypoint detection. Despite their effectiveness, these methods usually modify the original MLLM architecture and require training on large-scale datasets for specific downstream tasks, resulting in models specialized for those tasks. In contrast, VisHarness learns a generalizable policy for harnessing visual experts, enabling flexible application across diverse vision tasks.

\subsection{Tool-Augmented Visual Agents}
Tool-integrated reasoning alleviates the limitations of LLMs in precise computation and access to up-to-date information by invoking external tools such as code interpreters, calculators, and browsers during inference \cite{qian2025toolrl, wang2025actingless}. 
Beyond text-based tool use, recent works have begun to study tool-integrated agents with visual inputs. 
OPENTHINKIMG~\cite{su2025openthinkimg} studies tool-augmented visual question answering and uses RL to train MLLMs to adaptively invoke external vision tools. LLaVA-Plus~\cite{liu2024llava} augments multimodal assistants with a repository of vision and vision-language tools for diverse multimodal tasks. 
Another line of work develops MLLM-based agents that interact with promptable segmentation models~\cite{SegAgent, RSAgent, IBISAgent, SAM-Veteran}. SegAgent~\cite{SegAgent} emulates human annotation by having an MLLM provide positive and negative point prompts to a promptable segmentation model and iteratively refine segmentation masks through multi-turn interaction. RSAgent~\cite{RSAgent} supplements a promptable segmentation model with tools such as cropping and rotation, and uses SFT and RL to train an MLLM for multi-turn tool use. However, it still relies on the MLLM to directly generate point prompts for the segmentation model and selects only one target per image–text pair during training. These design settings limit its ability to handle dense multi-object scenes.
Unlike these works, VisHarness focuses on learning a generalizable policy to adaptively harness heterogeneous visual experts for diverse fundamental vision tasks under challenging conditions, including scenes with anywhere from zero to hundreds of target objects.

\section{VisHarness: Overall Design and Workflow}
\label{sec:method}

\subsection{Multi-Turn Visual Expert Harnessing}

\begin{figure}[t]
  \centering
  \includegraphics[width=0.95\textwidth]{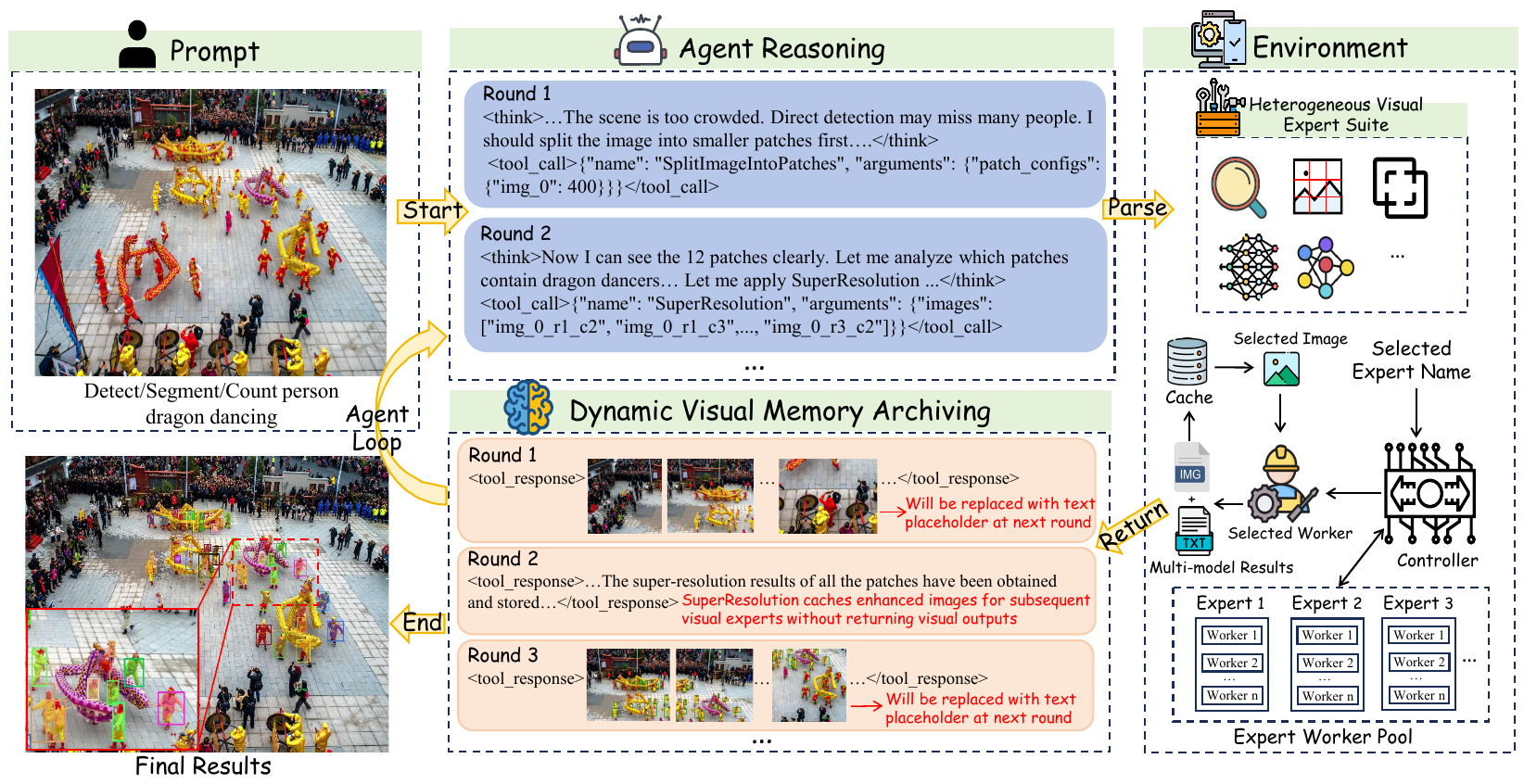}
  \caption{%
    \textbf{Overview of VisHarness.}
    VisHarness solves various complex fundamental vision tasks through multi-turn interactions with heterogeneous visual experts. At each turn, it selects an action based on the current memory. When a visual expert is invoked, the environment parses the expert name and arguments, and a controller dispatches the request to the least-loaded worker for that expert. After receiving the visualized and textual feedback, VisHarness dynamically archives outdated visual memory, incorporates the latest feedback into its memory, and proceeds to the next turn.}
  \label{fig:architecture}
\end{figure}

We propose VisHarness, a general-purpose visual agent that decouples high-level visual perception, reasoning, and decision-making from low-level visual task execution. It can effectively solve various complex visual tasks through multi-turn interactions with a set of heterogeneous visual experts. As shown in Fig. \ref{fig:architecture}, given an image-text pair $(I_i, T_i)$, $T_i$ specifies the objects to detect, segment, or count in $I_i$. The text can be a simple noun phrase (e.g., a class name), a complex attribute description (e.g., relative spatial relations, material, pose, or state), or an implicit query requiring reasoning to identify the target objects. The number of target objects can range from zero to over a hundred. 

VisHarness performs multi-turn reasoning over images and text, and finally produces the results ${(B_j, M_j)}_{j=1}^{\hat{n}}$, where $\hat{n}$ is the number of predicted target objects, and $B_j$ and $M_j$ denote the bounding box and corresponding mask of the $j$-th object, respectively. 
At the $t$-th turn, VisHarness $\pi_\theta$ selects an action $a_t=(\omega_t,\phi_t)\sim\pi_\theta(\cdot\mid\mathcal{M}_t)$ based on its current multimodal memory $\mathcal{M}_t$, where $\omega_t\in\Omega$ denotes the selected visual expert from the expert set $\Omega$, and $\phi_t\in\Phi_{\omega_t}$  denotes its invocation parameters. The environment then parses the action $a_t$ and executes the selected expert $\omega_t$ with parameters $\phi_t$, producing expert feedback $o_t=(v_t,r_t)\sim\mathcal{E}(\cdot\mid\omega_t,\phi_t)$, where $v_t$ and $r_t$ denote the visual and textual feedback, respectively.
 Then, the memory is updated by archiving earlier visual feedback and incorporating the current action and its corresponding feedback:
 \begin{equation}
  \mathcal{M}_{t+1}=\mathcal{U}(\mathcal{A}(\mathcal{M}_t),a_t,o_t),
  \label{eq:memory_update}
  \end{equation} where $\mathcal{A}(\cdot)$ denotes the visual archival function and $\mathcal{U}(\cdot)$ is the memory update function. Specifically, $\mathcal{A}(\cdot)$ replaces earlier visual feedback in the context memory with textual placeholders while preserving the textual history. This reduces the visual context overhead accumulated during multi-turn interactions. This process can be naturally formulated as a Markov decision process (MDP), where $\mathcal{M}_t$ serves as the state, $a_t=(\omega_t,\phi_t)$ serves as the action, and $\mathcal{M}_{t+1}$ is the next state.
Note that, to process multiple image-text pairs in parallel, each visual expert model is instantiated as multiple workers in the environment. When VisHarness requests to invoke a specific expert, a controller selects the worker with the lowest current load among all workers of that expert to execute the visual task.

\subsection{Heterogeneous Visual Expert Suite}
\label{sec:tools}
A sufficiently general and well-generalized visual expert suite is an important foundation for VisHarness to handle diverse and complex visual tasks. In our design, each expert performs a key atomic visual task. These experts have their own strengths and limitations, but under the unified coordination of VisHarness, they can complement one another and leverage their respective advantages, thereby solving complex visual tasks that are difficult for any single expert to handle alone.

\begin{figure}[t]
  \centering
  \includegraphics[width=\textwidth]{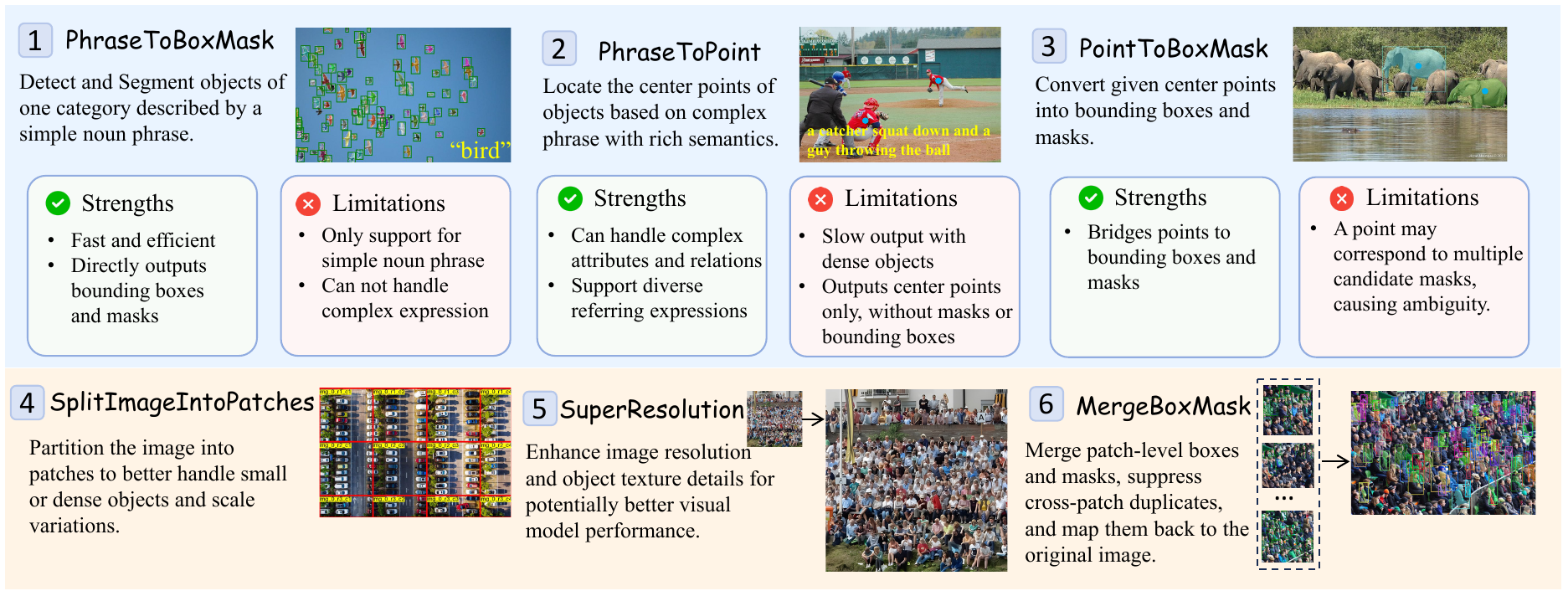}
  \caption{%
    The heterogeneous visual expert suite consists of six visual experts, including three fundamental visual models for executing specific vision tasks and three auxiliary visual tools for extending the capabilities of these models. Each visual expert plays a distinct role, while different experts can complement each other to jointly solve complex visual tasks that are difficult for any single expert to accomplish.}
  \label{fig:tools}
\end{figure}

As shown in Fig. \ref{fig:tools}, our expert suite consists of six experts, including three visual expert models and three auxiliary operation tools. The visual expert models are responsible for executing specific visual tasks, such as returning object bounding boxes, masks, or center points, while the auxiliary visual tools extend the capability range of these expert models to some extent. For example, in extremely dense scenes with small objects or scenarios with large scale variations, general-purpose visual expert models often struggle to handle the input images directly. In such cases, ``SplitImageIntoPatches'' divides the entire image into multiple patches, increasing the relative size of small objects within their respective patches. VisHarness can then actively select patches containing the target objects and apply appropriate vision expert models to process them. This overall strategy improves the accuracy of small-object detection and segmentation while enabling different subsequent processing strategies to be selected based on the content of each patch. Furthermore, the ``SuperResolution'' expert enhances the visual details of objects within a specified patch, potentially helping other vision expert models detect and segment blurry objects more accurately. Finally, the ``MergeBoxMask'' tool leverages the overlapping partitioning strategy of ``SplitImageIntoPatches'' to remove duplicate detections of the same object across different patches, producing complete and consistent final results.

\section{Learning to Harness Visual Experts}
\label{sec:training}
VisHarness is trained in two stages. First, we perform supervised fine-tuning (SFT) for cold start, enabling the model to generate visual expert calls in the required format. For on-policy reinforcement learning, we use Group Relative Policy Optimization (GRPO) \citep{deepseek2025r1} to train VisHarness in a live system equipped with visual expert models to further enhance its generalization ability. Due to space constraints, please refer to the supplementary material for details on SFT.
\subsection{GRPO with Dynamic Visual Memory Archiving}

GRPO requires sampling multiple trajectories for each input and optimizes the policy model by comparing their relative advantages within the same group. However, in VisHarness, multi-turn interactions with visual experts cause the returned visualized results to continuously accumulate in an uncontrollable manner, leading to substantial token overhead. 
To address this issue, we introduce \emph{Dynamic Visual Memory Archiving} during GRPO rollout. The key idea is to preserve only the most recent visual feedback, while dynamically archiving previous visual feedback into compact textual placeholders. In this way, VisHarness can still access the full textual history, while avoiding the continuous accumulation of visual tokens.

Given a training example $(I,T,y)$, where $(I,T)$ is an image-text pair and $y$ denotes the ground-truth visual annotation, we use the old policy model $\pi_{\theta_{\mathrm{old}}}$ to sample a group of $G$ trajectories:
\begin{equation}
    \{\tau_i\}_{i=1}^{G} \sim \pi_{\theta_{\mathrm{old}}}(\cdot \mid I,T),
\end{equation}
where the $i$-th trajectory is denoted as $\tau_i = \{(\mathcal{M}_{i,t}, a_{i,t},o_{i,t})\}_{t=1}^{T_i}$. Here, $a_{i,t}$ denotes the action generated by the policy at step $t$, and $o_{i,t}$ denotes the corresponding observation returned by the selected visual expert model. The action $a_{i,t}$ is sampled conditioned on the current memory $\mathcal{M}_{i,t}$, i.e. $a_{i,t} \sim \pi_{\theta_{\mathrm{old}}}
    \left(\cdot \mid \mathcal{M}_{i,t}\right)$.
The memory is updated dynamically during rollout using Eq.~\eqref{eq:memory_update}.
Concretely, before adding the latest observation, the archiving function $\mathcal{A}$ replaces earlier visual feedback with
textual placeholders while preserving the full textual history. The update function $\mathcal U$ then incorporates the current action $a_{i,t}$ and its observation $o_{i,t}=(v_{i,t},r_{i,t})$ into memory. Subsequent actions are thus conditioned on the latest
visual feedback together with the textual interaction history.

Since Dynamic Visual Memory Archiving replaces earlier visual feedback with textual placeholders at each turn, the contexts used across turns are not successive prefixes of a single fixed conversation sequence. To ensure context consistency between rollout and training in GRPO, we decompose each complete trajectory into turn-level training samples. Each sample contains the actual contextual memory used at that turn and the corresponding model response, with the training objective applied only to that response. This enables the model to learn to reason and act based on the corresponding context available at each turn. Specifically, after rolling out $G$ trajectories $ \{\tau_i\}_{i=1}^{G}$ of an example $(I,T,y)$, trajectory-level reward $r_i = \mathcal{R}(\hat{y}_i, y)$ is computed for each trajectory $\tau_i$, where $\hat{y}_i$ denotes the predicted visual results and $\mathcal{R}$ is the reward function, whose computation is detailed in the next subsection. Then, the group-relative advantage for the $i$-th trajectory is computed:
\begin{equation}
	\hat A_i
	=
	\frac{
		r_i-\mathrm{mean}\left(\{r_j\}_{j=1}^{G}\right)
	}{
		\mathrm{std}\left(\{r_j\}_{j=1}^{G}\right)
		+\epsilon_{\mathrm{adv}}
	}.
	\label{eq:group_relative_advantage}
\end{equation}
 Here, $\epsilon_{\mathrm{adv}}>0$ is a small constant for
 numerical stability.
 Since the reward function $\mathcal{R}$ evaluates the overall correctness of each complete trajectory, all turn-level training samples from a trajectory $\tau_i$ share its trajectory-level advantage $\hat{A}_i$. Each trajectory $\tau_i$ is decomposed into $T_i$ turn-level training samples, with $\hat{A}_i$ assigned to every response token in these samples. For each original image-text pair, the resulting turn-level training dataset $\mathcal{D}_{\mathrm{step}}$ contains a total of $\sum_{i=1}^{G} T_i$ training samples:
\begin{equation}
    \mathcal{D}_{\mathrm{step}}
    =
    \left\{ 
    	x_{i,t} = 
        \left(\mathcal{M}_{i,t}, a_{i,t}\right)
        \mid
        i=1,\dots,G,\;
        t=1,\dots,T_i
    \right\}.
\end{equation}
Let $a_{i,t,k}$ denote the $k$-th response token at turn $t$ of trajectory $\tau_i$, and let $L_{i,t}$ denote the response length.
Finally, the policy model is optimized over the turn-level training samples using the GRPO objective:
\begingroup
\small
\begin{equation}
	\resizebox{0.94\linewidth}{!}{$\displaystyle
		\mathcal L_{\mathrm{GRPO}}(\theta)
		=
		-\frac{1}{G}
		\sum_{i=1}^{G}
		\frac{1}{T_i}\sum_{t=1}^{T_i}
		\frac{1}{L_{i,t}}\sum_{k=1}^{L_{i,t}}
		\Big[
		\min\Big(
		\rho_{i,t,k}(\theta)\hat A_i,
		\operatorname{clip}\big(
		\rho_{i,t,k}(\theta),1-\epsilon,1+\epsilon
		\big)\hat A_i
		\Big)
		-\beta D_{\mathrm{KL}}^{i,t,k}
		(\pi_\theta\|\pi_{\mathrm{ref}})
		\Big].
	$}
	\label{eq:grpo_dvmc_objective}
\end{equation}
\endgroup
where
\begin{equation}
	\rho_{i,t,k}(\theta)
	=
	\frac{
		\pi_\theta\big(
		a_{i,t,k} \mid \mathcal M_{i,t},a_{i,t,<k}
		\big)
	}{
		\pi_{\theta_{\mathrm{old}}}\big(
		a_{i,t,k} \mid \mathcal M_{i,t},a_{i,t,<k}
		\big)
	},
\end{equation}
$\epsilon$ is the clipping coefficient, $\beta$ controls the strength of KL regularization, and $\pi_{\mathrm{ref}}$ denotes the fixed reference policy.
$D_{\mathrm{KL}}^{i,t,k}$ denotes the KL penalty estimated at response token $a_{i,t,k}$, conditioned on the memory $\mathcal M_{i,t}$ and response prefix $a_{i,t,<k}$.
By optimizing each action under its corresponding rollout memory using the trajectory-level advantage, this process enables GRPO training for multi-turn visual expert interaction while maintaining context consistency between rollout and training.

\subsection{Terminal Visual Rewards}
For image-text pairs whose visual ground truth is provided as point annotations, we design a soft reward function based on the relative counting error. For image-text pairs whose visual ground truth is provided as masks, we adopt an IoU-based soft reward function. Specifically, we define the unified soft reward as
\begin{equation}
r =
\begin{cases}
8\,\mathrm{IoU}(\hat{z}, z^\star)-3,
& \text{if the visual GT is provided as masks}, \\[4pt]
\max\!\left(-3,\ 5 - 8\dfrac{|\hat{n}-n^\star|}{n^\star}\right),
& \text{if the visual GT is point annotations},
\end{cases}
\label{eq:soft_reward}
\end{equation}
Here, $z^\star$ and $\hat z$ denote the ground-truth and predicted foreground masks, respectively. For multiple targets, their masks are merged by taking the union before computing IoU. $n^\star$ and $\hat n$ denote the ground-truth and predicted counts, respectively, with $n^\star>0$.

\section{Experiments}
\label{sec:experiments}

\subsection{Evaluation Setup}

\paragraph{Benchmarks.}
We evaluate VisHarness on four benchmarks spanning segmentation, detection, and counting, where the tasks further involve visual grounding conditioned on complex referring phrases, as well as dense- and small-object detection and counting. Specifically, gRefCOCO~\citep{liu2023gres} evaluates generalized referring expression segmentation under single-object, multi-object, and no-object queries. ReasonSeg~\citep{lai2024lisa} evaluates reasoning segmentation, requiring the model to infer the target objects from implicit textual queries before segmenting them. Dense200~\citep{jiang2026rexomni} evaluates dense small-object detection, while REC-8K~\citep{dai2024rec8k} evaluates referring-expression counting for dense objects described by complex phrases. Dense200 is held out from all SFT and GRPO training data to demonstrate the generalizability of the learned visual expert-harnessing policy. Please refer to the supplementary material for implementation details and evaluation metrics.

\begin{table*}[t]
\centering
\small
\caption{Generalized Referring Expression Segmentation on gRefCOCO. \textbf{Bold}: best.}
\label{tab:gres}
\renewcommand{\arraystretch}{1.0}
\setlength{\tabcolsep}{5.5pt}
\begin{tabular}{l|c|cc|cc|cc|c}
\toprule
\multirow{2}{*}{Method} & \multirow{2}{*}{\makecell{Training \\ Data Ratio}} & \multicolumn{2}{c|}{Validation Set} & \multicolumn{2}{c|}{Test Set A} & \multicolumn{2}{c|}{Test Set B} & \multirow{2}{*}{Avg.} \\
 &  & gIoU & cIoU & gIoU & cIoU & gIoU & cIoU &  \\ \midrule
\multicolumn{9}{c}{\textit{Traditional Specialized Segmentation Models}} \\
LTS~\cite{jing2021locate} (CVPR'2021) & 100\% & 52.70 & 52.30  & 62.64 & 61.87  & 50.42 & 49.96 & 54.98\\
VLT~\cite{ding2021vision} (ICCV'2021) & 100\% & 52.00 & 52.51  & 63.20 & 62.19  & 50.88 & 50.52 & 55.22 \\
CRIS~\cite{wang2022cris} (CVPR'2022) & 100\% & 56.27 & 55.34  & 63.42 & 63.82 & 51.79 & 51.04 & 56.95 \\
ReLA~\cite{liu2023gres} (CVPR'2023) & 100\% & \underline{63.60} & \underline{62.42}  & 70.03 & 69.26  & 61.02 & 59.88 & \underline{64.37} \\
LAVT~\cite{yang2022lavt} (CVPR'2022) & 100\% & 58.40 & 57.64  & 65.90 & 65.32  & 55.83 & 55.04 & 59.69 \\ \midrule
\multicolumn{9}{c}{\textit{Specialized Segmentation Models Based on MLLM}} \\
LISA(FT)~\citep{lai2024lisa} (CVPR'2024) & 100\% & 61.63 & 61.76 & 66.27 & 68.50 &  58.84 & 60.63 & 62.94 \\
GSVA(FT)~\cite{xia2024gsva} (CVPR'2024) & 100\% & \textbf{66.47} & \textbf{63.29}  & \textbf{71.08} & \underline{69.93}  & \underline{62.23} & 60.47 & \textbf{65.58} \\
LIRA-8B~\cite{LIRA} (ICCV'2025) & 0\% & 36.70 & 40.90 & 50.40 & 52.40 & 42.40 & 44.90 & 44.62 \\
SELF1E-8B~\cite{SELF1E} (CVPR'2026) & 0\% & 37.50 & 44.40 & 53.20 & 57.50 & 45.60 & 50.90 & 48.18 \\ \midrule
\multicolumn{9}{c}{\textit{Expert-Harnessing Visual Agents}} \\
Qwen3-VL-8B (w/ experts) & 0\% & 44.61 & 24.36 & 25.79 & 21.45 & 25.05 & 16.65 & 26.32\\
VisHarness-SFT (only SFT) & \textcolor{red}{0.76\%} & 51.45  & 57.04 & 67.40  & 68.13  & 60.91 & \underline{60.64} & 60.93   \\
VisHarness-RL (only RL) & \textcolor{red}{0.76\%} & 56.37 & 57.43 & 69.98 & 68.86 & 61.23 & 57.43 & 61.88 \\
VisHarness & \textcolor{red}{0.76\%} & 56.92  & 60.38  & \underline{70.53} & \textbf{70.63} & \textbf{63.27}  & \textbf{62.10}  & 63.97  \\
\bottomrule
\end{tabular}
\end{table*}

\subsection{Comparison with State-of-the-Art Methods}

\subsubsection{Generalized Referring Expression Segmentation}
\label{sec:gres}
The GRES task requires the model to segment specific targets in an image according to a given sentence. The target referred to by the sentence may be absent from the image, or may correspond to one or multiple object instances.
Table~\ref{tab:gres} presents the comparison results between our method and other methods specifically designed for referring expression segmentation. VisHarness achieves an average score of 63.97, only 1.61 and 0.40 points below GSVA(FT) and ReLA, respectively, while achieving the best cIoU on testA and the best gIoU and cIoU on testB. Both GSVA(FT) and ReLA use the full gRefCOCO training set, and GSVA(FT) additionally fine-tunes a pretrained MLLM using ten other segmentation datasets during mixed training~\cite{xia2024gsva}. In contrast, although VisHarness also builds on a pretrained MLLM and vision expert models, it achieves results comparable to or even better than those of specialized models using only about $0.76\%$ of the gRefCOCO training set. This indicates that our method can learn a generalizable vision expert harnessing policy from very limited data, substantially reducing reliance on task-specific training data while achieving strong performance.

Compared with Qwen3-VL-8B (w/ tools), VisHarness-SFT raises the average score from 26.32 to 60.93 (+34.61 points), showing that effective task execution requires learning an expert-harnessing policy rather than simply equipping a pretrained MLLM with visual experts. RL further improves all six metrics, raising the average to 63.97 and demonstrating its effectiveness in refining the policy beyond SFT. Without distillation from  larger models, VisHarness-RL outperforms VisHarness-SFT overall using only trajectory-level task rewards, further supporting the need for RL training.
\Needspace{4\baselineskip}
\subsubsection{Reasoning Segmentation}
\label{sec:reasonseg}

\begin{wraptable}[14]{r}{0.5\textwidth}
  \centering
  \caption{Comparison with state-of-the-art reasoning segmentation methods on ReasonSeg.}
  \label{tab:reasonseg}
  \setlength{\tabcolsep}{3.5pt}
  \resizebox{0.5\textwidth}{!}{
      \begin{tabular}{l|cc|cc|c}
      \toprule
      \multirow{2}{*}{Method} & \multicolumn{2}{c|}{Val} & \multicolumn{2}{c|}{Test} & \multirow{2}{*}{Avg.} \\
       & gIoU & cIoU & gIoU & cIoU &  \\ \midrule
      LISA~\citep{lai2024lisa} (CVPR'2024) & 53.6 & 52.3 & 48.7 & 48.8 & 50.9 \\
      SegLLM~\cite{wang2025segllm} (ICLR'2025) & 57.2 & 54.3 & 52.4 & 48.4 & 53.1 \\
      Seg-Zero~\cite{Seg-zero} (arXiv'2025) & 62.6 & 62.0 & 57.5 & 52.0 & 58.5 \\
      HiMTok-8B~\cite{HiMTok} (ICCV'2025) & 60.7 & \underline{67.0} & 60.8 & 66.2 & 63.7 \\
      Text4Seg++~\cite{Text4Seg++} (TPAMI'2026) & 59.1 & 49.5 & 57.1 & 52.1 & 54.5 \\
      SELF1E-8B~\cite{SELF1E} (CVPR'2026) & 65.9 & \textbf{69.7} & 65.7 & 67.0 & 67.1 \\
      SAM3-Agent~\cite{SAM3} (ICLR'2026) & 62.2 & 49.1 & 63.0 & 53.5 & 57.0\\ \midrule
      Qwen3-VL-8B (w/ experts) & 3.6 & 1.3 & 4.2  & 5.2 & 3.6 \\
      VisHarness-SFT (only SFT) & \underline{73.0} &  62.7  &  \underline{73.3} &  \underline{67.1} &  \underline{69.0}  \\
      VisHarness-RL (only RL) & \underline{73.0} & 62.6 & 71.6 & 61.2 & 67.1 \\
      VisHarness & \textbf{73.1}  & 62.1 & \textbf{73.8} & \textbf{71.8}  & \textbf{70.2} \\ \bottomrule
      \end{tabular}
  }
\end{wraptable}

In reasoning segmentation~\citep{lai2024lisa}, the target object
is not explicitly identified in the text but must instead be
inferred from the textual description. VisHarness naturally supports this task through its ``perception--reasoning--decision--execution'' capability: the MLLM first infers the target object from the text and then passes it as an argument to the visual expert for segmentation.
As shown in Table~\ref{tab:reasonseg}, VisHarness achieves
the best overall performance, with an average score of 70.2,
surpassing SELF1E-8B and SAM3-Agent (Qwen2.5-VL-7B)
by 3.1 and 13.2 points, respectively.
It also achieves the best results on three of the four
evaluation metrics.

Compared with Qwen3-VL-8B (w/ tools), VisHarness-SFT raises the average score from 3.6 to 69.0. The baseline frequently omits the required \texttt{MergeBoxMask} call for final submission, yielding empty predictions and highlighting the importance of learning a complete expert-harnessing policy through SFT. RL further raises the average to 70.2, with the largest gain of 4.7 points in test cIoU. The modest gain may reflect the task's focus on inferring targets from implicit queries, with a relatively fixed expert-calling strategy and limited room for expert coordination. This may also explain why VisHarness-RL, trained solely with task-outcome rewards, achieves a lower average score than VisHarness-SFT (67.1 vs.\ 69.0).

\Needspace{26\baselineskip}
\subsubsection{Dense and Tiny Object Detection}
\label{sec:detection}

\begingroup
\setlength{\intextsep}{0pt}
\begin{wraptable}[27]{r}{0.52\textwidth}
    \centering
    \captionsetup{position=top}
    \caption{Comparison on Dense200. All agents use Qwen3-VL-8B. Unreported results are denoted by --.}
    \label{tab:Dense200}
    \footnotesize
    \setlength{\tabcolsep}{2pt}
    \renewcommand{\arraystretch}{1.08}
    \resizebox{\linewidth}{!}{
    \begin{tabular}{@{}ll|ccc@{}}
    \toprule
    Method & Venue & \makecell{F1@IoU\\0.5} & \makecell{F1@IoU\\0.95} & \makecell{Mean\\F1} \\
    \midrule
    \multicolumn{5}{c}{\textit{Traditional Open-Set Detectors}} \\
    Grounding DINO~\citep{liu2024grounding} & ECCV'2024 & 36.9 & \textbf{19.7} & 33.1 \\
    \midrule
    \multicolumn{5}{c}{\textit{General-Purpose MLLMs}} \\
    DeepSeek-VL2-Tiny~\citep{wu2024deepseekvl2} & arXiv'2024 & 2.2 & 0.3 & 1.5 \\
    OVIS2.5-9B~\citep{lu2025ovis25} & arXiv'2025 & 14.0 & 0.0 & 5.1 \\
    OVIS2.5-2B~\citep{lu2025ovis25} & arXiv'2025 & 17.9 & 0.0 & 6.7 \\
    MiMo-VL~\citep{xiaomi2025mimovl} & arXiv'2025 & 29.7 & 0.4 & 15.9 \\
    Qwen2.5-VL-3B~\citep{bai2025qwen25vl} & arXiv'2025 & 0.8 & 0.1 & 0.5 \\
    Qwen2.5-VL-7B~\citep{bai2025qwen25vl} & arXiv'2025 & 1.1 & 0.1 & 0.6 \\
    Qwen3-VL-8B~\citep{bai2025qwen3vl} & arXiv'2025 & 15.90 & 2.41 & 11.44 \\
    Qwen3.5-9B~\citep{qwen3.5} & -- & 40.46 & 7.76 & 30.02 \\
    DeepSeek-VL2-Small~\citep{wu2024deepseekvl2} & arXiv'2024 & 16.0 & 3.9 & 12.7 \\
    SEED1.5-VL~\citep{guo2025seed15vl} & arXiv'2025 & 76.9 & 5.3 & 53.2 \\
    Moondream 3.1~\citep{moondream2026v31} & -- & 74.61 & -- & -- \\
    \midrule
    \multicolumn{5}{c}{\textit{Specialized Visual Models Based on MLLM}} \\
    Rex-Omni-SFT~\citep{jiang2026rexomni} & CVPR'2026 & 60.2 & 10.6 & 46.4 \\
    Rex-Omni~\citep{jiang2026rexomni} & CVPR'2026 & 78.4 & 10.3 & 58.3 \\
    LocateAnything~\citep{wang2026locateanything} & ECCV'2026 & 74.0 & \underline{18.5} & 58.7 \\
    SenseNova-Vision~\citep{han2026sensenovavision} & arXiv'2026 & -- & -- & 66.8 \\
    \midrule
    \multicolumn{5}{c}{\textit{Expert-Harnessing Visual Agents}} \\
    Qwen3-VL (w/ experts)~\citep{bai2025qwen3vl} & arXiv'2025 & 10.6 & 1.6 & 8.4 \\
    VisHarness-SFT (only SFT) & Ours & \textbf{83.9} & 16.8 & \textbf{68.7} \\
    VisHarness-RL (only RL) & Ours & 73.3 & 14.8 & 60.0 \\
    VisHarness & Ours & \underline{82.7} & 15.8 & \underline{67.5} \\
    \bottomrule
  \end{tabular}
    }
\end{wraptable}

Table~\ref{tab:Dense200} presents the dense small-object detection
results on the Dense200 dataset. We compare against the open-set
detector GroundingDINO~\citep{liu2024grounding}, general-purpose
MLLMs, and specialist detection models.
Notably, our training data do not include any Dense200 samples.
Nevertheless,
VisHarness-SFT and VisHarness achieve the best and second-best
mean F1 scores of 68.7 and 67.5, respectively.
VisHarness surpasses Rex-Omni and SenseNova-Vision
by 9.2 and 0.7 points in mean F1.
These strong results on a detection benchmark outside the
training tasks and datasets demonstrate the generalizability
of the learned expert-harnessing policy.

However, RL slightly reduces F1@IoU~0.5 from 83.9 to 82.7,
F1@IoU~0.95 from 16.8 to 15.8, and mean F1 from 68.7 to 67.5.
A likely explanation is the mismatch between the RL objective
and detection evaluation. RL uses no detection-task data.
For REC-8K, the RL supervision consists of object counts
derived from point annotations rather than bounding boxes.
The reward in Eq.~\ref{eq:soft_reward} compares predicted
and ground-truth counts without assessing box localization.
This misalignment may limit further gains in box quality. It may also explain why
VisHarness-RL falls well below VisHarness-SFT (60.0 vs.\ 68.7 mean F1).
\par\WFclear
\endgroup

\Needspace{4\baselineskip}
\subsubsection{Referring Expression Counting and Localization}
\label{sec:rec}

\noindent
\begin{minipage}[t]{0.46\textwidth}
  \vspace{0pt}
Table~\ref{tab:REC8K} shows that VisHarness achieves slightly lower counting
  and localization performance than GDino (FT) and GroundingREC, both
  fine-tuned on REC-8K. However, it also supports complex language-guided
  pixel-level segmentation, offering broader applicability. It outperforms
  Molmo2-4B, the model behind expert \texttt{PhraseToPoint}, in test MAE
  (8.91 vs.\ 10.04) and F1 (0.59 vs.\ 0.51), highlighting the benefit of
  coordinating heterogeneous experts over a single expert.
  As in other tasks, the untrained agent performs
  poorly, while RL improves SFT's test MAE from 10.34 to 8.91 and F1
  from 0.56 to 0.59.
  VisHarness-RL has slightly worse counting performance but better
  localization than VisHarness-SFT.
\end{minipage}\hfill
\begin{minipage}[t]{0.5\textwidth}
    \vspace{0pt}
    \centering
    \captionsetup{type=table}
    \caption{Comparison of specialized methods and general-purpose MLLMs on REC-8K. F1: point localization; FT: fine-tuning on REC-8K.}
    \label{tab:REC8K}
    \setlength{\tabcolsep}{2pt}
    \resizebox{\linewidth}{!}{
     \begin{tabular}{l | c c c | c c c}
\toprule
\multirow{2}{*}{Method} & \multicolumn{3}{c|}{Val set} & \multicolumn{3}{c}{Test set} \\
\cline{2-7}
 & MAE$\downarrow$ & RMSE$\downarrow$ & F1$\uparrow$ & MAE$\downarrow$ & RMSE$\downarrow$ & F1$\uparrow$ \\
\midrule
Mean              & 14.28 & 27.75 & -- & 13.75 & 25.91 & -- \\
ZSC (ResNet-50)~\cite{ZSOC} (CVPR'2023) & 14.84 & 31.30 & -- & 14.93 & 29.72 & -- \\
ZSC (Swin-T)~\cite{ZSOC} (CVPR'2023) & 12.96 & 26.74 & -- & 13.00 & 29.07 & -- \\
TFOC~\cite{shi2024training} (WACV'2024) & 16.08 & 31.61 & 0.12 & 17.27 & 32.68 & 0.11 \\
CounTX~\cite{amini2023open} (BMVC'2023) & 11.88 & 27.04 & -- & 11.84 & 25.62 & -- \\
CountGD~\cite{amini2024countgd} (NeurIPS'2024) & 9.51 & 22.91 & -- & 11.33 & 30.87 & -- \\
GDino~\cite{liu2024grounding} (ECCV'2024) & 11.77 & 28.60 & 0.34 & 11.71 & 26.97 & 0.35 \\
GDino (FT)~\cite{liu2024grounding} (ECCV'2024) & 9.03 & \underline{21.98} & \underline{0.65} & \underline{8.88} & \underline{21.95} & \underline{0.66} \\
GroundingREC~\cite{dai2024rec8k} (CVPR'2024) & \textbf{6.80} & \textbf{18.13} & \textbf{0.68} & \textbf{6.50} & \textbf{19.79} & \textbf{0.69} \\ \midrule
Qwen3-VL-8B~\cite{bai2025qwen3vl} (arXiv'2025) & 11.62 & 29.33 & 0.40 & 13.08 & 36.97 & 0.38 \\
Qwen3.5-9B \cite{qwen3.5}  & 9.78 & 24.90 & 0.45 & 10.18 & 31.44 & 0.47 \\
Molmo2-4B~\cite{Molmo2,triaridis2026crex} (CVPR'2026) & -- & -- & -- & 10.04 & 32.18 & 0.51 \\ \midrule
Qwen3-VL-8B (w/ experts) & 16.15 & 31.39 & 0.13 & 16.73 & 35.04 & 0.13 \\
VisHarness-SFT (only SFT) & 9.85 & 23.18 & 0.56 & 10.34 & 26.52 & 0.56 \\
VisHarness-RL (only RL) & 10.00 & 29.67 & 0.58 & 10.52 & 33.22 & 0.58 \\
VisHarness       & \underline{8.67} & 24.76 & 0.59 & 8.91 & 25.24 & 0.59 \\
\bottomrule
\end{tabular}
    }
\end{minipage}
\par

\Needspace{4\baselineskip}
\subsection{In-Depth Analysis}
\label{sec:ablation}

\subsubsection{Effect of Multi-Turn Interaction}
Fig.~\ref{fig:turn_ablation} presents interaction-turn ablations
on GRES and referring expression counting. We vary the maximum
number of interactions between the MLLM and visual experts from
2 to 12, reporting gIoU and cIoU for GRES, MAE and RMSE for
counting, and completion rates for both tasks. Low turn budgets
yield poor task performance and low completion rates, highlighting
the importance of multi-turn interaction. GRES performance and
completion rate largely stabilize after approximately six turns,
whereas referring expression counting requires approximately
eight interaction turns for its performance and completion
rate to stabilize.
\par\vspace{6pt}

\noindent
\begin{minipage}[t]{0.48\textwidth}
  \vspace{0pt}
  \centering
  \captionsetup{type=figure}
  \includegraphics[width=\linewidth]{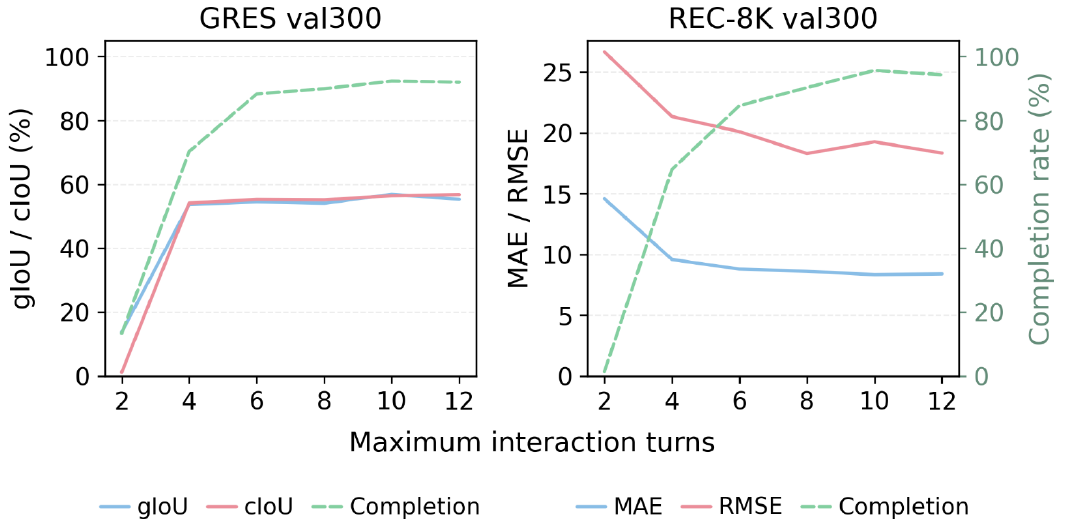}
  \caption{Effect of the number of interaction turns on performance and completion rate.}
  \label{fig:turn_ablation}
\end{minipage}\hfill
\begin{minipage}[t]{0.48\textwidth}
  \vspace{0pt}
  \centering
  \captionsetup{type=table}
  \caption{Effect of the Dynamic Visual Memory Archiving. Tokens denotes the average number of visual tokens per trajectory, and KV denotes peak active KV-cache memory per GPU (GiB).}
  \label{tab:archiving}
  \setlength{\tabcolsep}{3pt}
  \begin{tabular*}{\linewidth}{@{\extracolsep{\fill}}lrrrr@{}}
    \toprule
    \textit{GRES} & gIoU $\uparrow$ & cIoU $\uparrow$ & Tokens $\downarrow$ & KV $\downarrow$ \\
    w/o & \textbf{56.24} & 55.76 & 4,608.37 & 7.66 \\
    w/ & 55.82 & \textbf{56.32} & \textbf{1,080.70} & \textbf{5.05} \\
    \midrule
    \textit{REC-8K} & MAE $\downarrow$ & RMSE $\downarrow$ & Tokens $\downarrow$ & KV $\downarrow$ \\
    w/o & 7.97 & 16.27 & 15,253.19 & 22.03 \\
    w/ & \textbf{7.84} & \textbf{15.81} & \textbf{4,136.22} & \textbf{10.16} \\
    \bottomrule
  \end{tabular*}
\end{minipage}
\par

\subsubsection{Effect of Dynamic Visual Memory Archiving}
Table~\ref{tab:archiving} compares VisHarness with and without
Dynamic Visual Memory Archiving on subsets of the GRES and referring expression counting validation sets. Both settings achieve comparable task performance,
suggesting that replacing outdated visual feedback from experts
with textual placeholders reduces visual tokens while retaining
sufficient information for subsequent decision-making.
Specifically, archiving reduces average visual
tokens per trajectory by 76.5\% and 72.9\%, respectively, substantially
lowering the MLLM's context overhead and making both SFT and RL
training more memory efficient. After training, we deploy
VisHarness with vLLM for inference. Archiving also reduces
peak active KV-cache memory per GPU
by 34.1\% and 53.9\%, respectively.

\subsubsection{Visual Tools Usage Distribution Analysis}
Fig.~\ref{fig:tool_usage_dis} visualizes the distributions of visual
expert invocations across different models and tasks.
Fig.~\ref{fig:tool_usage_dis}(a) and Fig.~\ref{fig:tool_usage_dis}(b)
compare Qwen3-VL-8B and VisHarness on Dense200. VisHarness
increases the \texttt{MergeBoxMask} call ratio from 8.7\%
to 33.7\%. Since \texttt{MergeBoxMask} is required to submit
the final predictions, the baseline's frequent omission of
this call helps explain its poor performance across tasks.
Fig.~\ref{fig:tool_usage_dis}(b) and Fig.~\ref{fig:tool_usage_dis}(d)
compare VisHarness on Dense200 and gRefCOCO. On Dense200,
VisHarness primarily uses \texttt{PhraseToBoxMask} for detection
based on class names. On gRefCOCO, it relies more on
\texttt{PhraseToPoint} to localize targets described by complex
referring expressions before segmentation. This indicates that
VisHarness adapts its expert invocation pattern to task requirements.
Fig.~\ref{fig:tool_usage_dis}(c) and Fig.~\ref{fig:tool_usage_dis}(d)
compare VisHarness-SFT and VisHarness on gRefCOCO. RL reduces
the \texttt{PhraseToBoxMask} call ratio from 13.0\% to 8.6\%
and increases that of \texttt{PhraseToPoint} from 37.1\% to 39.3\%,
further reinforcing the expert invocation pattern needed for this task.

\subsubsection{Qualitative Analysis}
Fig.~\ref{fig:visual} shows two representative examples of multi-turn interaction. For GRES, VisHarness simultaneously locates the pink tie and pink phone with \texttt{PhraseToPoint}, then converts their centers into masks with \texttt{PointToBoxMask}. An unsuitable \texttt{mode} selects incorrect masks among multiple candidates for each point. VisHarness detects the error from visual feedback and switches \texttt{mode} from \texttt{area} to \texttt{confidence}, recovering the correct masks and demonstrating active error detection and correction. For referring expression counting, the query requests all people wearing black coats. VisHarness uses \texttt{SplitImageIntoPatches} to increase the relative size of these small, dense targets and \texttt{SuperResolution} to enhance texture details. It selects only patches containing targets for localization with \texttt{PhraseToPoint}, converts the points into masks and bounding boxes with \texttt{PointToBoxMask}, and removes duplicates across patches with \texttt{MergeBoxMask}. This demonstrates how partitioning simplifies dense small-object tasks while selective patch processing focuses on relevant regions. More visualizations and analyses are provided in Appendix~\ref{app:qualitative_results}.

\begingroup
\captionsetup{hypcap=false}
\begin{figure}[!htbp]
    \centering
    \includegraphics[width=\linewidth]{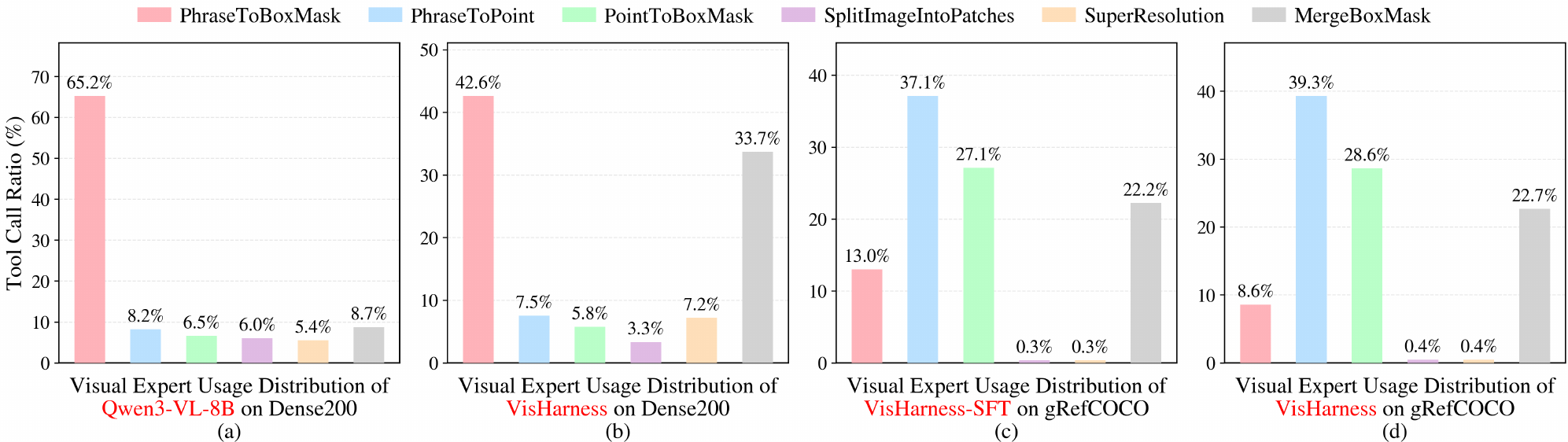}
    \caption{Distribution of visual expert calling by different model variants across different datasets.}
    \label{fig:tool_usage_dis}
    \par\vspace{6pt}
    \centering
    \includegraphics[width=\linewidth]{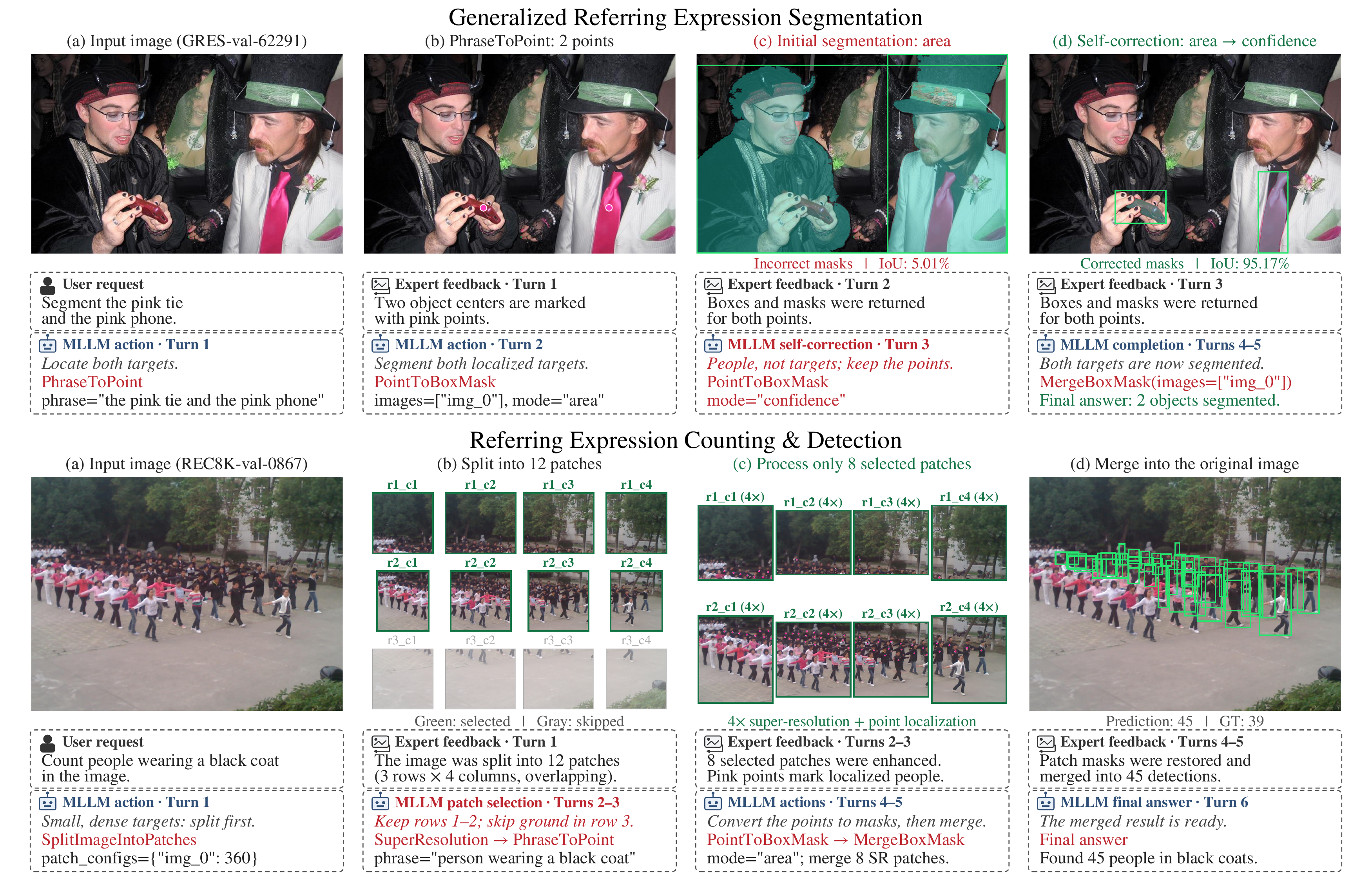}
    \caption{Multi-turn interaction visualization on two representative image-text pairs.}
    \label{fig:visual}
\end{figure}
\endgroup

\section{Conclusion}
\label{sec:conclusion}

In this paper, we presented \textbf{VisHarness}, a trainable visual agent that learns to harness heterogeneous visual experts for complex fundamental vision tasks. Instead of adapting an MLLM to solve a single visual task, VisHarness decouples high-level perception, reasoning, and decision-making from low-level visual execution, allowing the agent to preserve general vision-language reasoning ability while leveraging the precision of specialized visual models. Through multi-turn interactions with visual experts, VisHarness can adaptively select and coordinate heterogeneous visual experts at each turn, and self-correct to solve visual tasks involving complex language understanding and dense small-object perception. The dynamic visual memory archiving mechanism enables efficient reinforcement learning in a live environment with deployed visual experts without compromising contextual memory coherence or final task performance, further improving the generalizability of learned policy. Quantitative experiments show that VisHarness achieves comparable or superior performance to specialized models across tasks. Qualitative analysis further demonstrates that online reinforcement learning refines the distribution of expert invocations to better align with task-specific preferences, thereby improving task performance.

\begin{ack}
This work was supported by the National Natural Science Foundation of China (Grant Nos.~62276281 and U22A2095) and the Guangdong Basic and Applied Basic Research Foundation, China (Grant No.~2024A1515011882).
The authors declare no competing interests.
\end{ack}

\bibliographystyle{plainnat}
\bibliography{references}

\newpage
\appendix






\section{Additional Implementation Details}

\subsection{Evaluation Metrics}
For segmentation tasks, we report gIoU and cIoU. Given the predicted mask $\hat{M}_i$ and the ground-truth mask $M_i$ for the $i$-th sample, gIoU is computed as the average IoU over all $N$ evaluation samples:
\begin{equation}
	\mathrm{gIoU}
	=
	\frac{1}{N}
	\sum_{i=1}^{N}
	\frac{|\hat{M}_i \cap M_i|}{|\hat{M}_i \cup M_i|}.
\end{equation}
For samples with no target objects, the per-sample IoU is set to $1$ only when the model explicitly returns an empty final prediction, and to $0$ otherwise.
Multiple object masks are combined into a single foreground mask, and ignored pixels are excluded from evaluation.
cIoU measures the cumulative intersection-over-union over the whole dataset:
\begin{equation}
	\mathrm{cIoU}
	=
	\frac{\sum_{i=1}^{N} |\hat{M}_i \cap M_i|}
	{\sum_{i=1}^{N} |\hat{M}_i \cup M_i|}.
\end{equation}
If the cumulative union is zero, cIoU is defined as $0$.

For detection on Dense200, we report F1 scores at IoU thresholds of $0.5$ and $0.95$, denoted as $\mathrm{F1}_{0.5}$ and $\mathrm{F1}_{0.95}$, together with mF1.
For each threshold $\tau$, we process ground-truth boxes in order, matching each to the unused prediction with the highest IoU if this IoU is at least $\tau$.
Precision and recall are computed for each image-query pair and averaged separately over all $N$ samples:
\begin{equation}
	P_{\tau}
	=
	\frac{1}{N}
	\sum_{i=1}^{N}
	\frac{\mathrm{TP}_{i,\tau}}
	{\mathrm{TP}_{i,\tau}+\mathrm{FP}_{i,\tau}},
	\qquad
	R_{\tau}
	=
	\frac{1}{N}
	\sum_{i=1}^{N}
	\frac{\mathrm{TP}_{i,\tau}}
	{\mathrm{TP}_{i,\tau}+\mathrm{FN}_{i,\tau}},
\end{equation}
where $\mathrm{TP}_{i,\tau}$, $\mathrm{FP}_{i,\tau}$, and $\mathrm{FN}_{i,\tau}$ denote the numbers of true positives, false positives, and false negatives for sample $i$ at threshold $\tau$, respectively.
The F1 score is then computed as the harmonic mean of the averaged precision and recall:
\begin{equation}
	\mathrm{F1}_{\tau}
	=
	\frac{2P_{\tau}R_{\tau}}{P_{\tau}+R_{\tau}}.
\end{equation}
We define mF1 as the average F1 over ten IoU thresholds from $0.50$ to $0.95$ in increments of $0.05$:
\begin{equation}
	\mathrm{mF1}
	=
	\frac{1}{10}
	\sum_{k=0}^{9}
	\mathrm{F1}_{0.50+0.05k}.
\end{equation}
Any precision, recall, or F1 expression with a zero denominator is defined as $0$.

For counting tasks, we report MAE and RMSE.
Given the predicted count $\hat{n}_i$ and the ground-truth count $n_i$ for the $i$-th sample, the metrics are defined as
\begin{equation}
	\mathrm{MAE}
	=
	\frac{1}{N}
	\sum_{i=1}^{N}
	|\hat{n}_i-n_i|,
\end{equation}
and
\begin{equation}
	\mathrm{RMSE}
	=
	\sqrt{
		\frac{1}{N}
		\sum_{i=1}^{N}
		(\hat{n}_i-n_i)^2
	}.
\end{equation}

We additionally report localization F1.
We use the centers of predicted boxes as predicted points and normalize all point coordinates by the image width and height.
Predicted points and ground-truth points are matched using the Hungarian algorithm to minimize the total Euclidean distance.
For each sample with predictions, we sort the predicted boxes by area and select the middle box, using the upper middle box when the number of boxes is even.
The distance threshold is defined as
\begin{equation}
	\sigma_i
	=
	\frac{1}{2}
	\sqrt{(w_i^{\star})^2+(h_i^{\star})^2},
\end{equation}
where $w_i^{\star}$ and $h_i^{\star}$ are the normalized width and height of the selected box.
A matched pair is counted as a true positive only if its distance is strictly less than $\sigma_i$.
For each sample, $\mathrm{FP}_i=\hat{n}_i-\mathrm{TP}_i$ and $\mathrm{FN}_i=n_i-\mathrm{TP}_i$.
Localization F1 is computed by accumulating these counts across the dataset:
\begin{equation}
	\mathrm{F1}_{\mathrm{loc}}
	=
	\frac{2\sum_{i=1}^{N}\mathrm{TP}_i}
	{2\sum_{i=1}^{N}\mathrm{TP}_i
		+\sum_{i=1}^{N}\mathrm{FP}_i
		+\sum_{i=1}^{N}\mathrm{FN}_i}.
\end{equation}
Samples with no predictions have zero true positives.
Localization F1 is defined as $0$ when its denominator is zero.

\subsection{Cold-Start Data Generation}
We construct a seed pool of $4{,}499$ image--text pairs: $1{,}600$ from gRefCOCO~\cite{liu2023gres}, $2{,}660$ from REC-8K~\cite{dai2024rec8k}, and $239$ from ReasonSeg~\cite{lai2024lisa}. We first use Kimi-K2.5~\citep{kimi2026k25} to generate trajectories of multi-turn interactions with visual experts. For samples whose trajectories fail filtering, we regenerate trajectories using Qwen3.5-397B-A17B-FP8~\citep{qwen3.5}. The teachers invoke the deployed experts and continue interacting based on their actual visual and textual feedback. Ground-truth annotations are used only for subsequent filtering and are not provided as teacher inputs.

We retain only successfully completed trajectories. For gRefCOCO and ReasonSeg, the IoU between the predicted and ground-truth masks must be at least $0.7$; for REC-8K, the relative counting error must be at most $0.3$. Filtering yields $3{,}033$ complete trajectories: $1{,}226$ for gRefCOCO, $1{,}602$ for REC-8K, and $205$ for ReasonSeg.

SFT uses these filtered demonstrations, whereas RL uses all $4{,}499$ seed image--text pairs. All image--text pairs contributing to the SFT trajectories are included in the RL seed pool.

\subsection{Training Details}
We initialize VisHarness from Qwen3-VL-8B-Thinking~\citep{bai2025qwen3vl}. Only the agent is trained; the external visual experts remain frozen. Policy training uses $8$ NVIDIA A100 GPUs, while the visual experts are deployed on $4$ additional NVIDIA A100 GPUs. Each GPU has $80$~GB of memory. \texttt{SplitImageIntoPatches} and \texttt{MergeBoxMask} do not require GPUs.

\paragraph{SFT.}
Since Dynamic Visual Memory Archiving changes the context at each turn, we also decompose complete trajectories into turn-level training samples for SFT. We train on $13{,}323$ samples derived from the $3{,}033$ filtered trajectories. We perform full-parameter fine-tuning with Swift~\citep{zhao2025swift} for $1$ epoch, using a learning rate of $1\times10^{-5}$, a cosine learning-rate schedule, and a warmup ratio of $0.05$. SFT takes approximately $2.7$ hours.

\paragraph{RL.}
We implement multi-turn GRPO with \texttt{verl}~\cite{HybridFlow} and use vLLM for asynchronous rollouts. Training uses AdamW with a learning rate of $1\times10^{-6}$, a constant learning-rate schedule, no warmup, and $1$ epoch. For each update, we draw $8$ image--text pairs from the dataset and sample $8$ trajectories per pair. We retain a pair's trajectories for training only if the standard deviation of their rewards exceeds $0.2$. If any pairs are filtered out, we draw additional pairs and repeat rollout and filtering until $8$ pairs are retained. RL takes approximately $90.4$ hours, including periodic validation.

\paragraph{Validation.}
We use $300$ validation samples, selecting $100$ from the original validation split of each of gRefCOCO, REC-8K, and ReasonSeg. Samples are selected non-randomly to approximate the data distribution of the corresponding full validation split as closely as possible. Validation is performed before RL training and every $10$ RL training steps. For final evaluation, we use the RL checkpoint with the highest overall validation reward across the three tasks.

\subsection{Tool Deployment}
\texttt{PhraseToPoint} uses Molmo2-4B~\citep{Molmo2}; \texttt{PhraseToBoxMask} and \texttt{PointToBoxMask} use SAM3~\citep{SAM3}; and \texttt{SuperResolution} uses Real-ESRGAN~\citep{wang2021real}.

All experts are deployed as independent services and listen on specific ports. Each expert can be instantiated into multiple replicas to support parallel invocation by VisHarness, thereby improving inference efficiency. All expert instances are managed by a unified controller. When VisHarness needs to invoke an expert, it first sends a request to the controller. The controller then selects the instance with the lowest current workload among all replicas of the requested expert and returns its address, thereby achieving load balancing.

For full benchmark evaluation, we allow at most $12$ interaction turns. Each turn invokes at most one expert, and a single expert call may process multiple images.

\subsection{Cold-Start with SFT}
Cold-start SFT teaches the agent to generate valid expert calls before RL~\citep{hong2025deepeyesv2}, reducing format errors that could otherwise hinder effective exploration. We decompose the $3{,}033$ filtered trajectories into turn-level training samples. After excluding $14$ samples that exceed the visual-token budget, $13{,}323$ samples are used for SFT.

Each sample contains the actual context used at that turn and its corresponding teacher response. Loss is computed only on the current response tokens; historical model responses and expert feedback serve as conditioning inputs and are masked out of the loss. Let $\mathcal{D}_{\mathrm{SFT}}$ denote the retained turn-level samples, where $(\mathcal{M},a)$ contains the contextual memory and current response. The SFT objective is
\begin{equation}
    \mathcal{L}_{\mathrm{SFT}}
    =
    -\sum_{(\mathcal{M},a)\in\mathcal{D}_{\mathrm{SFT}}}
    \sum_{k=1}^{|a|}
    \log \pi_\theta(a_k \mid \mathcal{M},a_{<k}).
\end{equation}
Here, $a_k$ is the $k$-th token of the current response, and $a_{<k}$ denotes its preceding response tokens.

\newpage
\subsection{Prompt Templates}
\label{app:prompts} 
This section presents \texttt{DATA\_GENERATION\_SYSTEM\_PROMPT}, used for SFT data generation, and \texttt{TRAIN\_TEST\_SYSTEM\_PROMPT}, shared across SFT training, RL training, and inference. We also provide the descriptions and argument schemas of all experts in our implementation, followed by feedback prompts that guide the MLLM to correct errors or decide its next action after receiving expert outputs.

\subsubsection{System Prompt for SFT Data Generation}
This system prompt is used for SFT data generation and therefore provides detailed guidance on the rules to follow during trajectory generation. These rules guide the teacher models in generating SFT training data that are as accurate as possible.

\begin{bluepromptbox}{\texttt{DATA\_GENERATION\_SYSTEM\_PROMPT}}
You are an advanced vision-language AI agent whose goal is to accurately detect, segment, and count target objects in images specified by the user through queries by leveraging a series of specialized vision tools.

You MUST:
1. Carefully understand the user's query and analyze the provided image thoroughly, including object categories, attributes, and spatial relations.
2. Autonomously select the most suitable tool from the provided tool list at each step based on the image content and the user's query.
3. Carefully analyze the results returned by the tool, assess their accuracy and reliability, and decide the next action based on this assessment.

Image Naming Rules:
- The original input image is always named "img_0".
- If you split a image named "imageName" into patches, name patches as "imageName_r{row_index}_c{col_index}", where row_index and col_index start from 1.
- If you enhance the resolution of a image named "imageName", name the enhanced image as "imageName_4x".

Dynamic Conversation Context Management:
- Due to conversation context length limitations, only the most recent visualization returned by the tool are retained in the conversation history during each interaction.
The remaining images are replaced in conversation by a placeholder in the form "[System: Image 'xxx' has been archived to save memory.]". When you encounter such a placeholder, continue reasoning based on the descriptions in the conversation history.
- Images replaced by text placeholders are still retained in the system and can be accessed by their image name if needed for subsequent tool invocations.

Reminder:
1. The image may not contain the target object specified in the query. If you determine that the target object is absent, you can directly output the final answer. This scenario is intended to test the system's ability to handle cases where no target objects are present.
2. At each step, you can invoke only one tool and must not call multiple tools within one step.
3. For sentences that require reasoning, you should first infer the target object(s), and then plan which tool to invoke accordingly. For example, given an image containing a car and a bicycle, and the expression "the transportation vehicle that does not require electricity or gasoline," you should infer that the target object is the bicycle before invoking tools.
4. Whether the final result is obtained by directly invoking the vision tools on the original image, or by first applying SuperResolution or SplitImageIntoPatches, before output the final answer, you must first invoke MergeBoxMask to merge the results. Once the output of MergeBoxMask is confirmed to be correct, you can then output the final answer.
5. If the image contains the target object specified by the user, you must invoke at least one tool to ensure accuracy.
6. When you invoke multiple tools sequentially on the same image (either the same or different tools), the results from the previous tool call will be overwritten by subsequent results of the same type. For example, if bounding boxes and masks or points have already been obtained for an image, they will be replaced by the bounding boxes and masks or points returned by the next tool call.
7. The most important is that a tool's output may be inaccurate. You must evaluate tool output reliability before taking the next action, and if it is not, you may re-plan the subsequent steps accordingly.
8. All vision tools can accept multiple images of different sizes as input in a single invocation.
9. "SuperResolution" and "SplitImageIntoPatches" are key tools for improving performance. They should be used flexibly according to the density and size of target objects in the image to enhance the accuracy of detection, localization, and segmentation.
10. To ensure efficiency, the number of tool invocations should be minimized while maintaining accuracy.
11. Regardless of whether the task is detection, segmentation, or counting, you must obtain the bounding boxes and masks of the target objects by invoking tools, rather than providing a counting result based solely on your own estimation.
12. Keep your internal reasoning process thorough yet highly concise. Each step should be substantive but streamlined to avoid excessive length.
\end{bluepromptbox}

\subsubsection{System Prompt for SFT Training, RL Training, and Inference}
We use a more concise system prompt after SFT data generation, as the detailed rules are implicitly encoded in the SFT demonstrations.

\begin{orangepromptbox}{\texttt{TRAIN\_TEST\_SYSTEM\_PROMPT}}
You are an advanced vision-language AI agent whose goal is to accurately detect, segment, and count target objects in images specified by the user through queries by leveraging a series of specialized vision tools.

You MUST:
1. Carefully understand the user's query and analyze the provided image thoroughly, including object categories, attributes, and spatial relations.
2. Autonomously select the most suitable tool from the provided tool list at each step based on the image content and the user's query.
3. Carefully analyze the results returned by the tool, assess their accuracy and reliability, and decide the next action based on this assessment.

Output Format:
1. First provide the reasoning and thought process enclosed within <think></think> tags.
2. Then, either output a tool call directly if further action is needed or output a concise final answer enclosed in <answer></answer> if the task is complete.
3. The final answer should be a concise natural language summary of the tool's execution result, including confirmation of task completion, key statistics (e.g., detected classes and counts), and a brief reference to the generated visualization.

Image Naming Rules:
- The original input image is always named "img_0".
- If you split a image named "imageName" into patches, name patches as "imageName_r{row_index}_c{col_index}", where row_index and col_index start from 1.
- If you enhance the resolution of a image named "imageName", name the enhanced image as "imageName_4x".

Dynamic Conversation Context Management:
- Due to conversation context length limitations, only the most recent visualization returned by the tool are retained in the conversation history during each interaction.
The remaining images are replaced in conversation by a placeholder in the form "[System: Image 'xxx' has been archived to save memory.]". When you encounter such a placeholder, continue reasoning based on the descriptions in the conversation history.
- Images replaced by text placeholders are still retained in the system and can be accessed by their image name if needed for subsequent tool invocations.

Reminder:
1. At each step, you can invoke only one tool and must not call multiple tools within one step.
2. For sentences that require reasoning, you should first infer the target object(s), and then plan which tool to invoke accordingly.
3. When you invoke multiple tools sequentially on the same image (either the same or different tools), the results from the previous tool call will be overwritten by subsequent results of the same type.
4. Tools' output may be inaccurate. You must evaluate tool output reliability before taking the next action, and if it is not, you may re-plan the subsequent steps accordingly.
5. To ensure efficiency, the number of tool invocations should be minimized while maintaining accuracy.
6. Keep your internal reasoning process thorough yet highly concise. Each step should be substantive but streamlined to avoid excessive length.
\end{orangepromptbox}

\subsubsection{Expert Descriptions and Arguments}
These schemas describe the strengths, limitations, and suitable use cases of each expert, along with its calling interface and required arguments.

\begin{bluepromptbox}{\texttt{TOOLS\_LIST: PhraseToBoxMask}}
Return the bounding boxes and masks of all objects in the images that belong to the category described by the input simple noun phrase. This tool can only detect and segment objects of one category described by a simple noun phrase at a time and cannot detect and segment objects of multiple categories simultaneously.
Please note that the simple noun phrases here only consist of a noun or a single color + noun, such as "bus" and "yellow bus". Any phrase that includes implicit or explicit spatial information, material, posture, or any attributes other than single color does not meet the requirement for simple phrase in this tool, such as "the people in the second row", "the strawberries in the bowl", and "standing person."

Required arguments:
- images (array of strings): List of image names, e.g. ['img_0_r1_c1', 'img_0_r1_c2']
- phrase (string): Description of the target objects, e.g. 'yellow school bus'
\end{bluepromptbox}

\begin{bluepromptbox}{\texttt{TOOLS\_LIST: PhraseToPoint}}
Return the center points of all objects in the image matching the input complex phrases. This tool exhibits strong referring expression comprehension ability. It can localize multiple target objects of different categories contained in one input phrase at a time and can handle complex phrases that include implicit or explicit spatial information, material, posture, or any attributes beyond a single color.
You need to select an appropriate phrase to invoke this tool based on the image content and the user's query. An inappropriate phrase may affect the tool's performance. Therefore, if the returned results are inaccurate, you should adjust the phrase according to the image content and the user's query, and invoke the tool again to obtain more accurate results.
When there are many target objects of the same category need to be localized in the image, you may add "each" before the object, such as "each grape in the bowl." When there is only one target object, you may use "the," such as "the woman who is drinking water."
When multiple different objects in an image need to be localized, they must be described within a single phrase, and each object should be preceded by the definite article "the," for example: "the woman who is drinking water and the man who is looking to the right."
In some cases, this tool may produce severe hallucinations due to low input resolution, extremely dense objects, or other factors, resulting in neatly arranged but meaningless points. Therefore, before taking the next action, you should carefully evaluate whether this tool's output is correct.
Once the correct object centers are obtained via this tool, PointToBoxMask must be further invoked to produce the associated bounding boxes and masks, as they are required for the merging process.
Use this tool with caution. Invoke it only when the user prompt involves attributes beyond a single color, as this tool predicts coordinates sequentially and can be time-consuming (this may lead to a timeout error) when the image contains many objects or when many images need to be processed.
Overall, for phrases that exceed the scope of a noun or single color + noun, this tool should be given priority. Nevertheless, the number of target objects must be considered, since the tool may time out in cases of extremely dense object distributions.

Required arguments:
- images (array of strings): List of image names, e.g. ['img_0_r1_c1', 'img_0_r1_c2']
- phrase (string): Description of the object to locate, e.g. 'The goods on the third shelf'
\end{bluepromptbox}

\begin{bluepromptbox}{\texttt{TOOLS\_LIST: PointToBoxMask}}
Given the center points of objects in an image, it returns the corresponding bounding boxes and masks. The object center points used in this step must have been obtained in a previous step by calling the "PhraseToPoint" tool. When invoking this tool, you only need to provide the image name, and the system will retrieve the corresponding center points from the records.
This tool is primarily used in combination with Tool "PhraseToPoint" to detect and segment objects described by complex phrases.
Since a single point may correspond to multiple bounding boxes and masks, for example, when the point falls on a person's clothing, the returned bounding box and mask may correspond either to the clothing or to the entire person. To address this issue and improve the accuracy of the returned results, the tool provides two modes: "confidence" and "area". The "confidence" mode returns the bounding box and mask with the highest confidence for each point, while the "area" mode returns the bounding box and mask with the largest area for each point.
When the target objects are dense and small, and the local region around the point does not have a clear boundary from the whole object, the "confidence" mode should be used. When the target objects are relatively large and there is a clear boundary between the local region and the whole object (e.g., a clear and relatively large pedestrian where clothing and the whole person are clearly distinguishable), the "area" mode should be used. This parameter must be explicitly specified when invoking the tool. It is recommended to first try the "confidence" mode; if the returned result corresponds only to a local region and does not meet the requirement, then switch to the "area" mode.

Required arguments:
- images (array of strings): List of image names, e.g. ['img_0_r1_c1', 'img_0_r1_c2']
- mode (string; one of "confidence", "area"): Mode to determine bounding box, either 'confidence' or 'area'.
\end{bluepromptbox}

\begin{bluepromptbox}{\texttt{TOOLS\_LIST: SplitImageIntoPatches}}
Partition the input image into equally sized patches with a certain degree of overlap (to facilitate merging detection and segmentation results).
This tool is particularly effective when the target objects are small, or when there exist significant scale variations across the image.
For example, in dense crowd scenes, human heads closer to the camera appear much larger than those farther away.
By partitioning the image into uniform patches, each patch can be processed independently: patches containing dense small objects can be super-resolved, while patches with sparse and large objects can be directly handled by vision tools.
Patch partitioning should take into account the size, spatial distribution and density of target objects.
This tool can also isolate targets in specific regions by first partitioning the image into patches and then selecting relevant patches for detection, localization or segmentation, such as pedestrians in the second row or those crossing at the lower-left crosswalk.
For a given image, you need to determine an appropriate positive integer as the patch size (the width and height of the patch are equal). If the patch size is not divisible by this integer, the partitioning tool will return some patches whose width or height differs from the specified value.
The patch size is selected to balance detection accuracy and computational efficiency, avoiding redundant processing introduced by excessively small patches.

Required arguments:
- patch_configs (object mapping image names to integer values): A dictionary where each key is a string representing an image name (e.g., 'img_0_r1_c1'), and each value is an integer specifying the patch_size used for that image during partition.
Note that you need to choose an appropriate integer as the patch_size. Since the resulting patches are square (with equal width and height), only a single value is required to specify the side length, .e.g. {'img_0_r1_c1': 320, 'img_0_r1_c2': 240}
Value description: An integer that specifies the side length of square patches when partitioning the given image.
\end{bluepromptbox}

\begin{bluepromptbox}{\texttt{TOOLS\_LIST: SuperResolution}}
Increase the resolution of the given image by a factor of 4 to improve the visibility of small objects within it.
This tool is particularly useful for image containing small and densely packed objects that are difficult to discern at their original resolution.
Whether to apply super-resolution to the original image mainly depends on whether the target objects are too small to be detected by the vision tool, rather than only on object density. Even in dense scenes, if the objects are sufficiently large, the vision tool can be applied directly.

Required arguments:
- images (array of strings): List of image names, e.g. ['img_0_r1_c1', 'img_0_r1_c2']
\end{bluepromptbox}

\begin{bluepromptbox}{\texttt{TOOLS\_LIST: MergeBoxMask}}
Once you are certain that the entire object detection and segmentation pipeline has been completed, you can invoke this tool to obtain final results.
This tool first maps the bounding boxes and masks from all selected images back to the coordinate system of the original image (img_0). It then removes duplicate detections and merges fragmented detections belonging to the same object to produce the final bounding boxes and masks.
When invoking this tool, you must specify in the parameter list which patches contain valid bounding boxes and masks. During the trial-and-error process, some patches may produce redundant or invalid results. Therefore, you need to actively select only those patches that contribute valid bounding boxes and masks to the final result for merging.
Note that the patches you specify for merging the final result must have already been obtained the bounding boxes and masks of the target objects.
Whether the final result is obtained by directly invoking the vision tool on the original image, or by first applying SuperResolution or SplitImageIntoPatches, once you determine that the entire process has been completed, you must invoke this tool to generate the final result.

Required arguments:
- images (array of strings): List of image names of patchs that contain valid bounding boxes and masks to the final results, e.g. ['img_0_r1_c1', 'img_0_r1_c2_4x']
\end{bluepromptbox}

\Needspace{6\baselineskip}
\subsubsection{Expert Feedback Prompts}
After an expert returns its results, these prompts guide the MLLM to actively check for errors and decide how to proceed before taking the next action. The expert-specific feedback prompts are used only during SFT data generation. For SFT training, RL training, and inference, we retain only the general feedback prompt \texttt{VISION\_TOOL\_PROMPT}.

\begin{orangepromptbox}{\texttt{VISION\_TOOL\_PROMPT}}
Based on the returned text and visualized results, carefully assess the reliability of the tool outputs for each image. Check whether the reported object count matches the expected targets, and inspect the visualization to ensure that each point is centered on the corresponding target object or that each target object is tightly enclosed by its corresponding bounding box. If the result is inaccurate, adjust the tool parameters or revise the tool-use strategy and try again.
\end{orangepromptbox}

\begin{orangepromptbox}{\texttt{PHRASE\_TO\_BOXMASK\_PROMPT}}
If the bounding boxes shown in the visualization returned by PhraseToBoxMask are incorrect (e.g., some target objects are missed or non-target objects are falsely detected), first reassess whether the phrase used to invoke PhraseToBoxMask is too complex for the tool's capabilities. If the phrase contains complex attributes or relations, consider using PhraseToPoint followed by PointToBoxMask instead. If the target objects are too small or blurry to be detected reliably, consider using SplitImageIntoPatches and SuperResolution as appropriate before invoking the vision tool again.
\end{orangepromptbox}

\begin{orangepromptbox}{\texttt{PHRASE\_TO\_POINT\_PROMPT}}
If the results returned by PhraseToPoint are incorrect (e.g., the target objects are not localized, or low image resolution or overly blurry target objects result in neatly arranged hallucinated points), you can refine the phrase used to invoke PhraseToPoint to make it more appropriate or specific, or use SplitImageIntoPatches and SuperResolution as appropriate to improve the visibility of the target objects before invoking PhraseToPoint again. If all target object center points have already been correctly obtained via PhraseToPoint, PointToBoxMask must still be invoked to generate the bounding boxes and masks for all objects before invoking MergeBoxMask to generate the final result.
\end{orangepromptbox}

\begin{orangepromptbox}{\texttt{POINT\_TO\_BOXMASK\_PROMPT}}
If the bounding boxes shown in the visualization returned by PointToBoxMask are inappropriate (e.g., too large and exceeding the target object, or too small and covering only a local part of the object), you can adjust the "mode" parameter and invoke PointToBoxMask again.
\end{orangepromptbox}

\begin{orangepromptbox}{\texttt{SPLIT\_PROMPT}}
Based on the overall partitioning scheme and the local details of each patch, determine whether each patch contains the target objects. For patches containing target objects, decide whether to directly invoke the vision tool or to apply SuperResolution.
\end{orangepromptbox}

\begin{orangepromptbox}{\texttt{SR\_PROMPT}}
Based on the super-resolved images, decide whether to directly invoke the vision tool or to further partition them into patches.
\end{orangepromptbox}

\clearpage
\section{Additional Qualitative Results}
\label{app:qualitative_results}
\begin{figure}[!htbp]
    \centering
    \includegraphics[width=\linewidth]{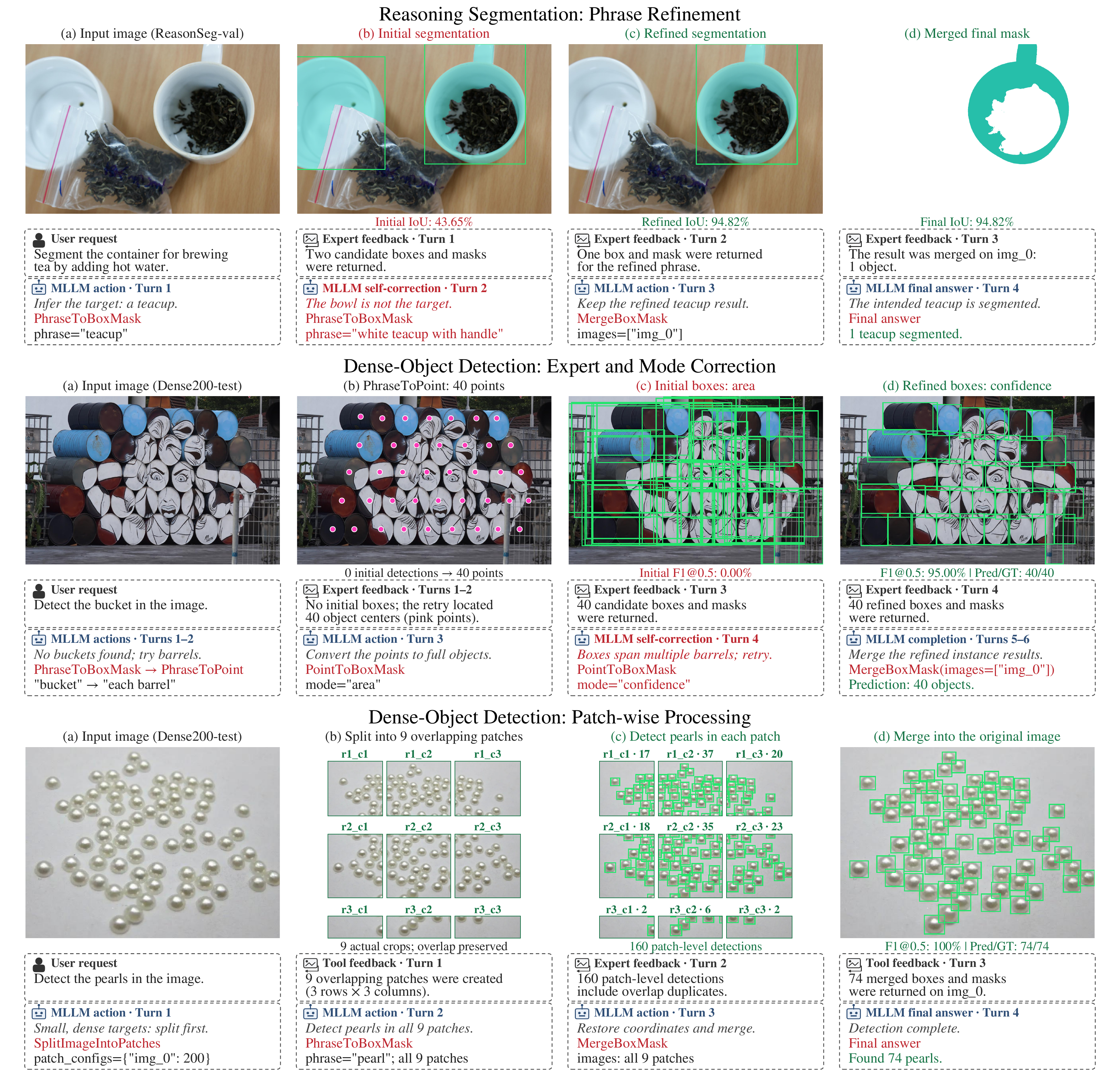}
    \caption{Additional multi-turn visual expert interactions. The three rows illustrate phrase refinement for reasoning segmentation, expert switching and parameter correction for dense-object detection, and patch-wise processing of small objects, respectively.}
    \label{fig:supplementary_visual_cases}
\end{figure}

Fig.~\ref{fig:supplementary_visual_cases} presents three additional examples of VisHarness adapting its visual expert calls based on intermediate feedback.

\noindent\textbf{Phrase refinement from visual feedback.}
In the first example, the implicit query refers to the cup containing tea leaves. VisHarness initially invokes \texttt{PhraseToBoxMask} with ``teacup,'' which also segments an empty container. After identifying this false positive in the visualization, it revises the phrase to ``white teacup with handle,'' using a distinguishing attribute inferred from the image rather than specified in the query. This isolates the intended cup and improves IoU from 43.65\% to 94.82\%, illustrating its ability to discover an alternative solution through visual reasoning.

\noindent\textbf{Expert switching and parameter correction.}
In the second example, \texttt{PhraseToBoxMask} returns no detections for ``bucket.'' VisHarness switches to \texttt{PhraseToPoint} with ``each barrel'' and obtains all 40 object centers. It then invokes \texttt{PointToBoxMask} with \texttt{mode="area"}, but observes that the resulting boxes span multiple barrels. Changing the parameter to \texttt{mode="confidence"} produces appropriate instance boxes and masks, increasing F1 at IoU 0.5 from 0.00\% to 95.00\%. This example shows that VisHarness can recover from expert failures by adjusting both expert selection and invocation parameters.

\Needspace{9\baselineskip}
\noindent\textbf{Patch-wise processing of small objects.}
In the third example, VisHarness first creates nine overlapping patches with \texttt{SplitImageIntoPatches}, increasing the relative size of the small pearls within each input. It applies \texttt{PhraseToBoxMask} to each patch with the phrase ``pearl,'' obtaining 160 patch-level detections, including overlap duplicates. Finally, \texttt{MergeBoxMask} maps the results back to the original image and removes duplicates, yielding 74 objects, matching the ground-truth count, with 100\% F1 at IoU 0.5. This demonstrates how coordinating image partitioning, expert execution, and result merging enables accurate detection of dense small objects.

\section{Limitations.}
\label{app:limitations}
Despite its strong performance across diverse visual tasks, VisHarness still has several limitations.
First, VisHarness relies on the coverage and reliability of the available visual experts. For a given image-query pair, if all candidate experts fail to correctly detect, localize, or segment the target objects, the agent cannot recover the correct result through high-level reasoning alone, and the task is likely to fail.
Second, VisHarness relies heavily on the MLLM's visual perception for error correction. If the MLLM fails to detect subtle errors in the visualizations returned by the experts, it cannot correct them, potentially leading to incorrect final results.
Third, deploying heterogeneous visual experts introduces additional computational and system overhead. Unlike a single end-to-end model, VisHarness requires multiple expert models to be deployed as callable services, which increases GPU memory usage, inference latency, and engineering complexity.

\section{Broader Impacts}
VisHarness may have positive impacts by improving the flexibility and accessibility of general-purpose visual understanding systems. By learning to harness heterogeneous visual experts, it can reduce the need to design separate task-specific models for detection, segmentation, and counting, and may benefit applications such as visual data analysis, assistive perception, and efficient annotation. This expert-harnessing paradigm may also benefit embodied agents. For example, an agent could invoke visual experts to detect, localize, and segment target objects to support manipulation tasks such as grasping. However, the same capability may also be misused in privacy-sensitive scenarios, such as automated surveillance or large-scale monitoring. In addition, incorrect tool outputs or erroneous agent decisions may lead to unreliable conclusions when the system is deployed in safety-critical settings. Therefore, practical deployment should include human oversight, careful validation of tool outputs, and appropriate restrictions for privacy-sensitive applications.
\newpage

\end{document}